\documentclass[10pt,twocolumn,letterpaper]{article}

\usepackage{iccv}
\usepackage{times}
\usepackage{epsfig}
\usepackage{graphicx}
\usepackage{amsmath}
\usepackage{amssymb}

\usepackage{float}
\usepackage{multirow}
\usepackage{footnote}
\usepackage{marvosym}

\usepackage[
pagebackref=true,
breaklinks=true,
letterpaper=true,
colorlinks,
bookmarks=false
]{hyperref}

\iccvfinalcopy 

\def\iccvPaperID{2385} 

\ificcvfinal\fi
\begin{document}

\title{SSAP: Single-Shot Instance Segmentation With Affinity Pyramid}

\author{
Naiyu~Gao$^{1,2}$,
Yanhu~Shan$^{3}$,
Yupei~Wang$^{1,2}$,
Xin~Zhao$^{1,2}$\thanks{Corresponding author},
Yinan~Yu$^{3}$,
Ming~Yang$^{3}$,
Kaiqi~Huang$^{1,2,4}$
\\
$^{1}$
CRISE, Institute of Automation, Chinese Academy of Sciences\\
$^{2}$
University of Chinese Academy of Sciences
$^{3}$
Horizon Robotics, Inc\\
$^{4}$
CAS Center for Excellence in Brain Science and Intelligence Technology\\
 {\tt\small \{gaonaiyu2017,wangyupei2014\}@ia.ac.cn,\{xzhao,kaiqi.huang\}@nlpr.ia.ac.cn}\\
 {\tt\small\{yanhu.shan,yinan.yu\}@horizon.ai, m-yang4@u.northwestern.edu
}}

\maketitle
\ificcvfinal\thispagestyle{empty}\fi

\begin{abstract}
Recently, proposal-free instance segmentation has received increasing attention due to its concise and efficient pipeline. Generally, proposal-free methods generate instance-agnostic semantic segmentation labels and instance-aware features to group pixels into different object instances. However, previous methods mostly employ separate modules for these two sub-tasks and require multiple passes for inference. We argue that treating these two sub-tasks separately is suboptimal. In fact, employing multiple separate modules significantly reduces the potential for application. The mutual benefits between the two complementary sub-tasks are also unexplored. To this end, this work proposes a single-shot proposal-free instance segmentation method that requires only one single pass for prediction. Our method is based on a pixel-pair affinity pyramid, which computes the probability that two pixels belong to the same instance in a hierarchical manner. The affinity pyramid can also be jointly learned with the semantic class labeling and achieve mutual benefits. Moreover, incorporating with the learned affinity pyramid, a novel cascaded graph partition module is presented to sequentially generate instances from coarse to fine. Unlike previous time-consuming graph partition methods, this module achieves $5\times$ speedup and 9\% relative improvement on Average-Precision (AP). Our approach achieves state-of-the-art results on the challenging Cityscapes dataset.
\end{abstract}

\section{Introduction}
The rapid development of Convolutional networks \cite{lecun1998gradient-based,krizhevsky2012imagenet} has revolutionized various vision tasks, enabling us to move towards more fine-grained understanding of images. Instead of classic bounding-box level object detection \cite{girshick2014rich,girshick2015fast,ren2015faster,liu2016ssd:,redmon2017yolo9000,dai2016r} or class-level semantic segmentation \cite{long2015fully,chen2018deeplabv2:,zhao2017pyramid}, instance segmentation provides in-depth understanding by segmenting all objects and distinguishing different object instances. Researchers are thus showing increasing interests in instance segmentation recently.

Current state-of-the-art solutions to this challenging problem can be classified into the \emph{proposal-based} and \emph{proposal-free} approaches~ \cite{liang2015proposal,Kirillov_2017_CVPR,Liu_2018_ECCV}. The proposal-based approaches regard it as an extension to the classic object detection task \cite{ren2015faster,liu2016ssd:,redmon2017yolo9000,dai2016r}. After localizing each object with a bounding box, a foreground mask is predicted within each bounding box proposal. However, the performances of these proposal-based methods are highly limited by the quality of the bounding box predictions and the two-stage pipeline also limits the speed of the systems. By contrast, the proposal-free approach has the advantage of its simple and efficient design. This work also focuses on the proposal-free paradigm.

\begin{figure}[t]
\begin{center}
\includegraphics[width=.98\linewidth]{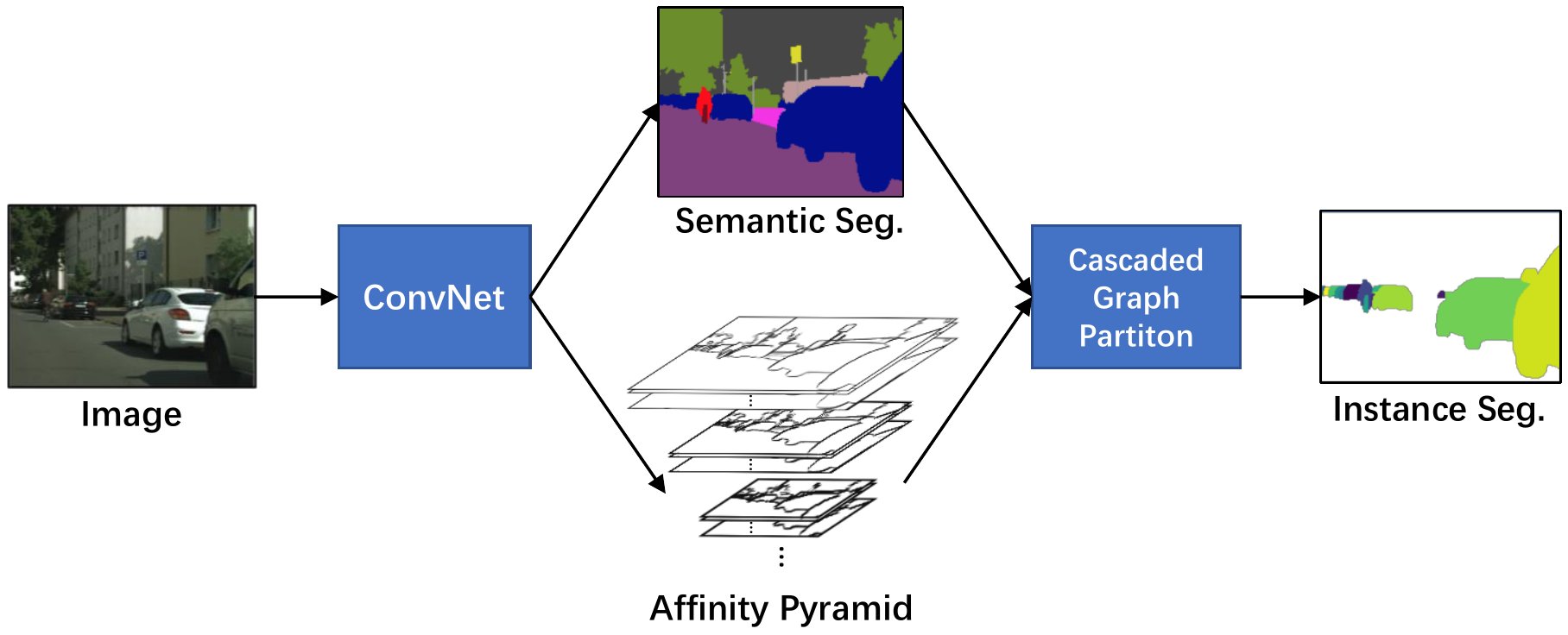}
\end{center}
   \vspace{-.0em}
   \caption{Overview of the proposed method. The per-pixel semantic class and pixel-pair affinities are generated with a single pass of a fully-convolutional network. The final instance segmentation result is then derived from these predictions by the proposed cascaded graph partition module.}
   \vspace{-.0em}
\label{fig:overview}
\end{figure}
The proposal-free methods mostly start by producing instance-agnostic pixel-level semantic class labels \cite{long2015fully,chen2018deeplabv2:,chen2017rethinking,zhao2017pyramid}, followed by clustering them into different object instances with particularly designed instance-aware features. However, previous methods mainly treat the two sub-processes as two separate stages and employ multiple modules, which is suboptimal. In fact, the mutual benefits between the two sub-tasks can be exploited, which will further improve the performance of instance segmentation. Moreover, employing multiple modules may result in additional computational costs for real-world applications.

To cope with the above issues, this work proposes a single-shot proposal-free instance segmentation method, which jointly learns the pixel-level semantic class segmentation and object instance differentiating in a unified model with a single backbone network, as shown in Fig. ~\ref{fig:overview}.
Specifically, for distinguishing different object instances, an affinity pyramid is proposed, which can be jointly learned with the labeling of semantic classes.
The pixel-pair affinity computes the probability that two pixels belong to the same instance.
In this work, the short-range affinities for pixels close to each other are derived with dense small learning windows.
Simultaneously, the long-range affinities for pixels distant from each other are also required to group objects with large scales or nonadjacent parts. 
Instead of enlarging the windows, the multi-range affinities are decoupled and long-range affinities are sparsely derived from instance maps with lower resolutions. 
After that, we propose learning the affinity pyramid at multiple scales along the hierarchy of an U-shape network, where the short-range and long-range affinities are effectively learned from the feature levels with higher and lower resolutions respectively.
Experiments in Table~\ref{table:ablation_feature} show that the pixel-level semantic segmentation and pixel-pair affinity pyramid based grouping are indeed mutually benefited from the proposed joint learning scheme. The overall instance segmentation is thus further improved. 

Then, in order to utilize the cues about global context reasoning, this work employs a graph partition method \cite{keuper2015efficient} to derive instances from the learned affinities. 
Unlike previous time-consuming methods, a cascaded graph partition module is presented to incorporate the graph partition process with the hierarchical manner of the affinity pyramid and finally provides both acceleration and performance improvements.
Concretely, with the learned pixel-pair affinity pyramid, a graph is constructed by regarding each pixel as a node and transforming affinities into the edge scores. Graph partition is then employed from higher-level lower-resolution layers to lower-level higher-resolution layers progressively. Instance segmentation predictions from lower resolutions produce confident proposals, which significantly reduce node numbers at higher resolutions. Thus the whole process is accelerated.

The main contributions of this paper are as follows:

\begin{itemize}
\vspace{-0.1em}\item A novel instance-aware pixel-pair affinity pyramid is proposed to distinguish instances, which can be jointly learned with the pixel-level labeling of semantic class. The mutual benefits between the two sub-tasks are explored by encouraging bidirectional interactions, which further boosts instance segmentation.
\vspace{-0.1em}\item A single-shot, proposal-free instance segmentation method is proposed, based on the proposed affinity pyramid. Unlike most previous methods, our approach requires only one single pass to generate instances.  
On the challenging Cityscapes dataset, our method achieves state of art with 37.3\% AP (val) / 32.7\% (test) and 61.1\% PQ (val).
\vspace{-0.1em}\item Incorporating with the hierarchical manner of the affinity pyramid, a novel cascaded graph partition module is proposed to gradually segment an image into instances from coarse to fine. Compared with the non-cascaded way, this module achieves $5\times$ speedup and 9\% relative improvement on AP.
\end{itemize}

\section{Related Work}
\label{section:instancereview}
\subsection{Instance Segmentation}
Existing approaches on instance segmentation could be divided into two paradigms: proposal-based methods and proposal-free methods.

Proposal-based methods recognize object instances with bounding boxes that generated with detectors \cite{ren2015faster,liu2016ssd:,dai2016r}.
MNC \cite{dai2016instance-aware} decomposes instance segmentation into a cascade of sub-tasks, including box localization, mask refinement and instance classification.
Another work \cite{Arnab_2017_CVPR,Li_2018_ECCV} combines the predictions of detection and semantic segmentation with a CRFasRNN \cite{zheng2015conditional} to generate instances.
FICS \cite{li2017fully} develops the position sensitive score map \cite{dai2016instance-sensitive}.
Mask R-CNN \cite{He_2017_ICCV} extends Faster R-CNN \cite{ren2015faster} by adding a segmentation mask predicting branch on each Region of Interest (RoI). Following works extend Mask R-CNN by modifying feature layers \cite{Liu_2018_CVPR} or the mask prediction head \cite{Chen_2018_CVPR}.

Proposal-free methods mainly solve instance segmentation based on the success of semantic segmentation \cite{chen2018deeplabv2:,zhao2017pyramid,chen2017rethinking}. The segmentation based methods learn instance-aware features and use corresponding grouping methods to cluster pixels into instances.
DWT \cite{Bai_2017_CVPR} learns boundary-aware energy for each pixel followed by watershed transform.
Several methods \cite{discriminative_loss_2017_cvprw,deep_metric,Neven_2019_CVPR} adopt instance level embeddings to differentiate instances.
SGN \cite{Liu_2017_ICCV} sequentially groups instances with three sub-networks.
Recurrent Neural Networks (RNNs) is adopted in several approaches \cite{romeraparedes2016recurrent,ren2017end-to-end} to generate one instance mask at each time.
Graph based algorithm \cite{keuper2015efficient} is also utilized for post-processing \cite{Levinkov_2017_CVPR,Kirillov_2017_CVPR}, which segments an image into instances with global reasoning. However, the graph based algorithm is usually time-consuming.
To speed up, Levinkov \etal \cite{Levinkov_2017_CVPR} down-sample the outputs before the graph optimization while Kirillov \etal \cite{Kirillov_2017_CVPR} only derive edges for adjacent neighbors. They all accelerate at the expense of performance.
Recently, Yang \etal \cite{arxiv_2019_yang_deeperlab} propose a single-shot image parser that achieves a balance between accuracy and efficiency.
\subsection{Pixel-Pair Affinity}
The concept of learning pixel-pair affinity has been developed in many previous works ~\cite{liu2017learning,ke2018adaptive,Ahn_2018_CVPR,bertasius2017convolutional,maire2016affinity} to facilitate semantic segmentations during training or post-processing.
Recently, Liu \etal \cite{Liu_2018_ECCV} propose learning instance-aware affinity and grouping pixels into instances with agglomerative hierarchical clustering. Our approach also utilizes instance-aware affinity to distinguish object instances, but both the ways to derive affinities and group pixels are significantly different. Importantly, Liu \etal \cite{Liu_2018_ECCV} employ two models and require multiple passes for the RoIs generated from semantic segmentation results. Instead, our approach is single-shot, which requires only one single pass to generate the final instance segmentation result.

\section{Proposed Approach}
This work proposes a single-shot proposal-free instance segmentation model based on the jointly learned semantic segmentation and pixel-pair affinity pyramid, which are equipped with a cascaded graph partition module to differentiate object instances.
As shown in Fig.~\ref{fig:pipeline}, our model consists of two parts: (a) a unified network to learn the semantic segmentation and affinity pyramid with a single backbone network, and (b) a cascaded graph partition module to sequentially generate multi-scale instance predictions using the jointly learned affinity pyramid and semantic segmentation.
In this section, the affinity pyramid is firstly explained at Subsection~\ref{section:AP}, then the cascaded graph partition module is described at Subsection~\ref{section:CGP}.

\subsection{Affinity Pyramid}
\label{section:AP}
\begin{figure}[t]
\begin{center}
  \includegraphics[width=0.95\linewidth]{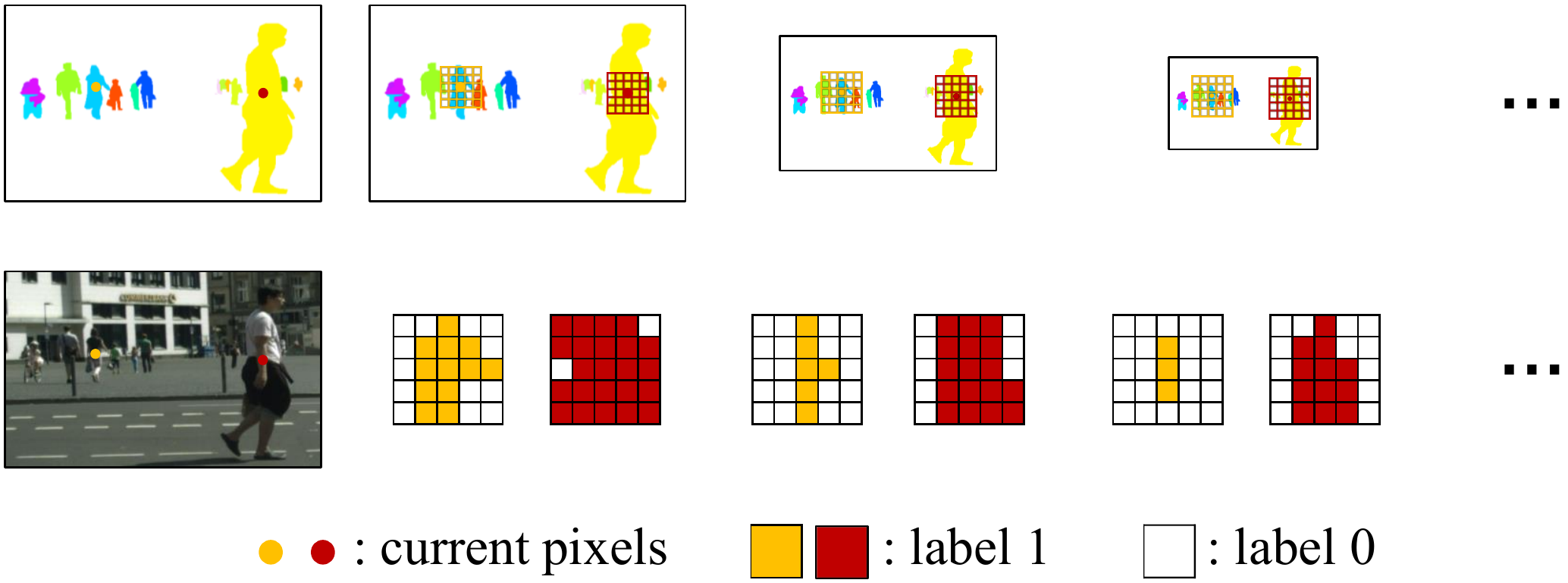}
\end{center}
\caption{Illustration of affinity pyramid. Pixel-pair affinity specifies whether two pixels belong to the same instance or not. For each current pixel, the affinities to neighboring pixels within a small $r\times r$ window (here, $r=5$) are predicted.
The short-range and long-range affinities are decoupled and derived from instance maps with higher and lower resolutions respectively.
In practice, ground truth affinity is set to 1 if two pixels are from the same instance, otherwise 0. Best viewed in color and zoom.}
\label{fig:AffinityPyramid}
\end{figure}

With the instance-agnostic semantic segmentation, grouping pixels into individual object instance is critical for instance segmentation. This work proposes distinguishing different object instances based on the instance-aware pixel-pair affinity, which specifies whether two pixels belong to the same instance or not.
As shown in the second column of Fig.~\ref{fig:AffinityPyramid}, for each pixel, the short-range affinities to neighboring pixels within a small $r \times r$ window are learned. In this way, a $r^2 \times h \times w$ affinity response map is presented.
For training, the average L2 loss is calculated with the $r^2$ predicted affinities for each pixel:
\begin{equation}
loss(a,y)=\frac{1}{r^2}\sum_{j=1}^{r^2}\big{(}y^j-a^j\big{)}^2,
\label{eq:al2}
\end{equation}
where $a=[a^1,a^2,\dots,a^{r^2}]$. $a^j$ is the predicted affinity between the current pixel and the $j{\text{-th}}$ pixel in its affinity window, representing the probability that two pixels belong to the same instance. The sigmoid activation is used to let $a^j\in (0,1)$. Here, $y=[y^1,y^2,\dots,y^{r^2}]$ and $y^j$ represents the ground truth affinity for $a^j$. $y^j$ is set to 1 if two pixels are from the same instance, 0 if two pixels are from different instances.
Importantly, the training data generated in this way is unbalanced. 
Specifically, the ground truth affinities are mostly with all $1$ as most pixels are at the inner-regions of instances. To this end, $80\%$ pixels with all $1$ ground truth affinities are randomly dropped during training.
Additionally, we set 3 times loss for pixels belonging to object instances.

\begin{figure*}[htbp]
\begin{center}
\includegraphics[width=0.91\linewidth]{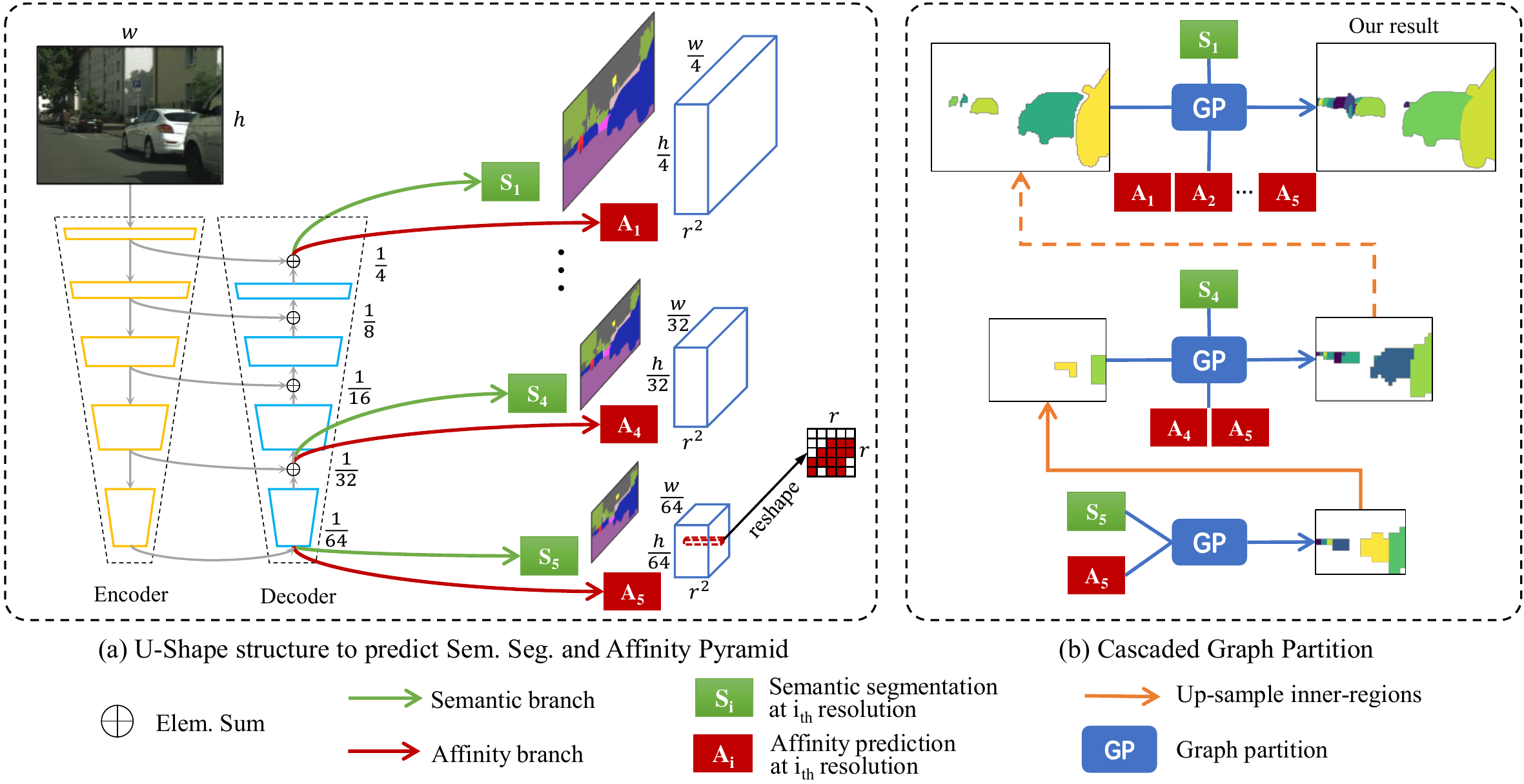}
\end{center}
   \vspace{-0.1em}
   \caption{Our instance segmentation model consists of two parts: 
   (a) a unified U-shape framework that jointly learns the semantic segmentation and affinity pyramid. The affinity pyramid is constructed by learning multi-range affinities from feature levels with different resolutions separately.
   (b) a cascaded graph partition module that utilizes the jointly learned affinity pyramid and semantic segmentation to progressively refine instance predictions starting from the deepest layer. Instance predictions in the lower-level layers with higher resolution are guided by the instance proposals generated from the deeper layers with lower resolution. Best viewed in color and zoom.}
\label{fig:pipeline}
\end{figure*}

Moreover, apart from the short-range affinities above, the long-range affinities are also required to handle objects of larger scales or nonadjacent object parts. A simple solution is to utilize a large affinity window size. 
However, besides the cost of GPU memories, a large affinity window would inevitably conflict with the semantic segmentation during training, which severely hinders the joint learning of the two sub-tasks. As shown in experiments (see Table~\ref{table:windowsize}), jointly learning the short-range affinities with semantic segmentation obtains mutual benefits for the two tasks. 
However, the long-range affinities are obviously more difficult to be jointly learned with the pixel-level semantic class labeling.
Similar observation is also captured by Ke \etal \cite{Ke_2018_ECCV}.

Instead of enlarging the affinity window, we propose to learn multi-scale affinities as an affinity pyramid, where the short-range and long-range affinities are decoupled and the latter is sparsely derived from instance maps with lower resolutions.
More concretely, as shown in Fig.~\ref{fig:AffinityPyramid}, the long-range affinities are achieved with the same small affinity window at the lower resolutions. Note that the window sizes can be different, however they are fixed in this work for simplicity.
In this way, the $5\times 5$ windows from the $\frac{1}{64}$ resolution can produce affinities between pixels at most 128 pixel-distance.
With the constructed affinity pyramid, the finer short-range and coarser long-range affinities are learned from the higher and lower resolutions, respectively. Consequently, multi-scale instance predictions are generated by affinities under corresponding resolutions. 
As shown in Fig.~\ref{fig:AffinityPyramid}, the predictions of larger instances are proposed by the lower resolution affinities, and are further detailed by higher resolution affinities. Meanwhile, although the smaller instances have too weak responses to be proposed at lower resolutions, they can be generated by the affinities with higher resolutions.

After that, the affinity pyramid can be easily learned by adding affinity branches in parallel with the existing branches for semantic segmentation along the hierarchy of the decoder network. As shown in Fig.~\ref{fig:pipeline} (a), affinities are predicted under $\{\frac{1}{4}, \frac{1}{8}, \frac{1}{16}, \frac{1}{32}, \frac{1}{64}\}$ resolutions of the original image. 
In this way, the short-range and long-range affinities can be effectively learned at different feature levels in the feature pyramid of the U-shape architecture. The formed affinity pyramid can thus be jointly learned with the semantic segmentation in a unified model, resulting in mutual benefits.

\subsection{Cascaded Graph Partition}
\label{section:CGP}
With the jointly learned semantic segmentation and affinity pyramid, a graph-based partition mechanism is employed in this work to differentiate object instances.
In particular, incorporating with the hierarchical manner of the affinity pyramid, a cascaded graph partition module is presented. This module sequentially generates instances with multiple scales, guided by the cues encoded in the deeper-level layers of the affinity pyramid.

\noindent\textbf{Graph Partition }
With the learned pixel-pair affinity pyramid, an undirected graph $G=(V,E)$ is constructed, where $V$ is the set of pixels and $E\subseteq V^2$ is the set of pixel-pairs within affinity windows. $e_{u,v}\in E$ represents the edge between the pixels $\{u,v\}$.
Furthermore, $a_{u,v}, a_{v,u}\in(0,1)$ are the affinities for pixels $\{u,v\}$, which are predicted at pixels $u$ and $v$, respectively.
The average affinity $\alpha_{u,v}$ is then calculated and transformed into the score $w_{u,v}$ of edge $e_{u,v}$ by:
\vspace{0.2em}
\begin{align}
\label{eq:average}
\alpha_{u,v}&=(a_{u,v}+a_{v,u})/2,\\
\label{eq:weight}
w_{u,v}&=\log(\frac{\alpha_{u,v}}{1-\alpha_{u,v}}).
\vspace{0.2em}
\end{align}
As the affinities predict how likely two pixels belong to the same instance, the average affinities higher than $0.5$ are transformed into positive and negative otherwise.
In this way, instance segmentation is transformed into a graph partition problem \cite{chopra1993the} and can be addressed by solving the following optimization problem \cite{keuper2015efficient}:
\vspace{0.2em}
\begin{align}
\label{eq:object}
   \min_{y \in \{0,1\}}&\sum_{e\in E}w_{e}y_{e},\\
\label{eq:constraince}
   s.t.\, \forall C \: \forall e^{'}\in C:&\sum_{e\in C \setminus \{e^{'}\}} y_e\ge y_{e^{'}}.
\end{align}
\vspace{0.2em}
Here, $y_e=y_{u,v}\in\{0,1\}$ is a binary variable and $y_{u,v}=1$ represents nodes $u$ and $v$ belong to different partitions. $C$ is the set of all cycles of the graph $G$.
The objective in formulation~\ref{eq:object} is about to maximize the total score of the selected edges, and the inequality~\ref{eq:constraince} constrains each feasible solution representing a partition.
A search-based algorithm \cite{keuper2015efficient} is developed to solve the optimization problem.
However, when this algorithm is employed to segment instances, the inference time is not only long but also rises significantly \emph{w.r.t.} the number of nodes, which brings potential problem for real-world applications.

\noindent\textbf{Cascade Scheme }
The sizes of instances in the Cityscapes dataset are various significantly. For large instances, the pixels are mostly at the inner-regions which cost long inference time although are easy for segmentation.
Motivated by this observation, a cascaded strategy is developed to incorporate the graph partition mechanism with the hierarchical manner of the affinity pyramid.
As shown in Fig.~\ref{fig:pipeline} (b), the graph partition is firstly utilized on a low resolution where it has fewer pixels and requires a short running time for graph partition.
Although only coarse segments for large instances are generated,  the inner-regions for these segments are still reliable.
In this case, these inner-regions can be up-sampled and regarded as proposals for the higher resolution. 
At the higher resolution, the pixels in each proposal are combined to generate a node and the remaining pixels are each treated as a node. 
To construct a graph with these nodes, the edge score $w_{t_i,t_j}$ between nodes $t_i$ and $t_j$ is calculated by adding all pixel-pair edge scores between the two nodes:
$w_{t_i,t_j}=\sum_{u\in t_i,v\in t_j}{w_{u,v}}$. 
In this way, the proposals for instance predictions are progressively refined.
Because the number of nodes decreases significantly at each step, the entire graph partition is accelerated.

\begin{figure}[t]
\begin{center}
\includegraphics[width=.95\linewidth]{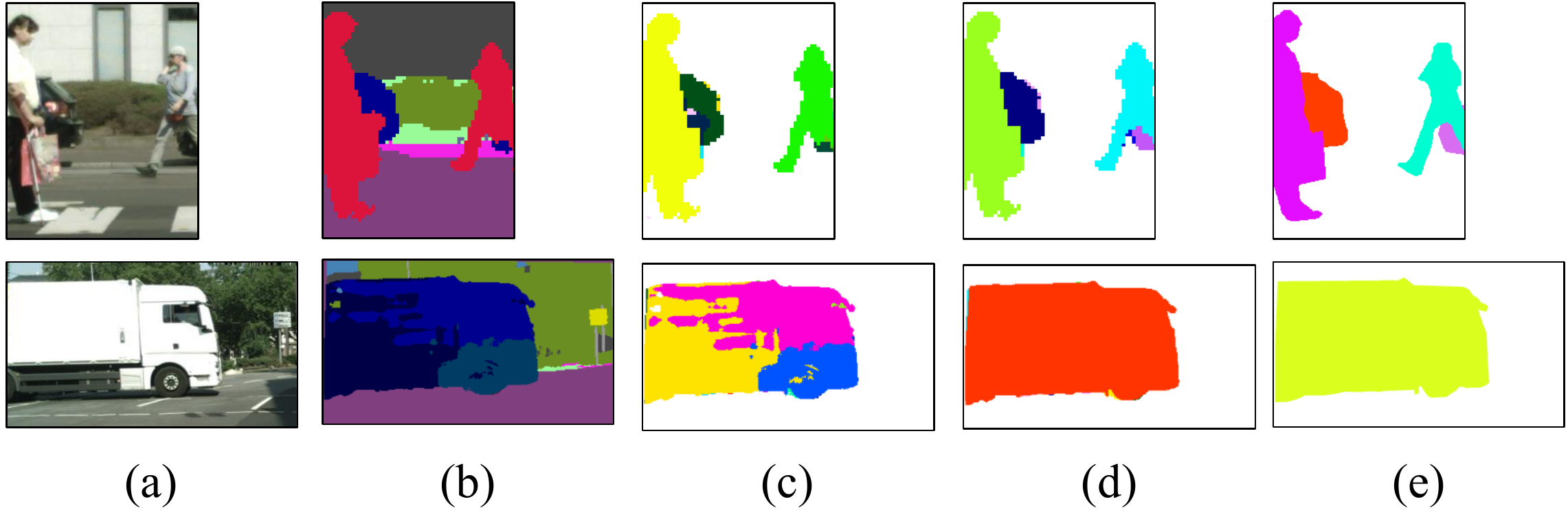}
\end{center}
\caption{Influence of segmentation refinement (SR). (a) Input image. (b) Semantic segmentation. (c) Instance segmentation without SR (d) Instance segmentation with SR. (e) Ground truth. SR significantly improves the errors in instance segmentation which are caused by the semantic segmentation failures. Best viewed in color and zoom.}
\label{fig:parsing_refine}
\end{figure}
\noindent\textbf{Segmentation Refinement }
In previous steps, the partition is made within each class to speed up. At this step, the cues from both semantic segmentation and affinity branches are integrated to segmentation instances from all the pixels which are classified as foreground. In practice, the average affinity $\alpha_{u,v}$ for pixels $\{u,v\}$ is refined to $\alpha^{'}_{u,v}$ by:
\begin{equation}
 \alpha^{'}_{u,v}=\alpha_{u,v}*\exp[-\mathrm{D_{JS}}(s_u\|s_v)],
\label{eq:score}
\end{equation}
\begin{equation}
\mathrm{D_{JS}}(P\|Q)=\frac{1}{2}\Big{[}\mathrm{D_{KL}}\big{(}P\|\frac{P+Q}{2}\big{)}+\mathrm{D_{KL}}\big{(}Q\|\frac{P+Q}{2}\big{)}\Big{]},
\label{eq:js}
\end{equation}
\begin{equation}
   \mathrm{D_{KL}}(P\|Q)=\sum_{i}P_i\log{\frac{P_i}{Q_i}}.
   \label{eq:kl}
\end{equation}
Here, $s_u=[s_u^1,s_u^2,\dots,s_u^c]$ and $s_v=[s_v^1,s_v^2,\dots,s_v^c]$ are the semantic segmentation scores for $c$ object classes at the pixel $u$ and $v$, which represent the classification possibility distributions on the $c$ object classes. The distance between the two distributions can be measured with the popular Jensen-Shannon divergence, as described in Eq.~\ref{eq:js}-\ref{eq:kl}.
After the refinement for the initial affinities, graph partition is conducted for all the foreground pixels at the $\frac{1}{4}$ resolution.
By combining the information from semantic segmentation and affinity branches, the errors in instance segmentation which are caused by the semantic segmentation failures are significantly improved, as shown in Fig.~\ref{fig:parsing_refine}.
 
Finally, the class label for each instance is obtained by voting among all pixels based on semantic segmentation labels.
Following DWT \cite{Bai_2017_CVPR}, small instances are removed and semantic scores from semantic segmentation are used to rank predictions.

\section{Experiments}
\noindent\textbf{Dataset }
Our model is evaluated on the challenging urban street scenes dataset Cityscapes \cite{cordts2016the}. In this dataset, each image has a high resolution of 1,024$\times$2,048 pixels. There are 5,000 images with high quality dense pixel annotations and 20,000 images with coarse annotations. Note that only the fine annotated dataset is used to train our model.
Cityscapes benchmark evaluates 8 classes for instance segmentation. Together with another 11 background classes, 19 classes are evaluated for semantic segmentation. 

\noindent\textbf{Metrics }
The main metric for evaluation is Average-Precision (AP), which is calculated by averaging the precisions under IoU (Intersection over Union) thresholds from 0.50 to 0.95 at the step of 0.05. Our result is also reported with three sub-metrics from Cityscapes: AP50\%, AP100m and AP50m. They are calculated at 0.5 IoU threshold or only for objects within specific distances. 

This paper also evaluates the results with a new metric Panoptic Quality (PQ) \cite{panoptic}, which is further divided into Segmentation Quality (SQ) and Recognition Quality (RQ) to measure recognition and segmentation performances respectively. The formulation PQ is defined as:
\begin{equation}
\small
\mathrm{PQ}=\underbrace{\frac{\sum_{p,g\in TP}{\mathrm{IoU}(p,g)}}{|TP|}}_{\text{Segmentation Quality (SQ)}}\times\underbrace{\frac{|TP|}{|TP|+\frac{1}{2}|FP|+\frac{1}{2}|FN|}}_{\text{Recognition Quality (RQ)}},
\end{equation}
where $p$ and $g$ are the predicted and ground truth segments, while $TP$, $FP$ and $FN$ represent matched pairs of segments, unmatched predicted segments and unmatched ground truth segments respectively.
Moreover, both countable objects (thing) and uncountable regions (stuff) are evaluated in PQ and are separately reported with PQ$^{\text{Th}}$ and PQ$^{\text{St}}$. As the stuff is not concerned in this work, only PQ and PQ$^{\text{Th}}$ is reported.

\begin{table}[t]
\begin{center}
\footnotesize
\begin{tabular}{cc|ccc}
\hline
$\alpha$&$\lambda$&AP (\%)&PQ$^{\text{Th}}$ (\%)&PQ (\%)\\
\hline
$0.0003$	&diff.	 &27.5 	&45.0&54.6\\
$0.001$		&diff.	 &29.5 	&48.0&55.8\\
$0.003$		&diff.	 &\textbf{31.5}	&\textbf{49.2}&\textbf{56.6}\\
$0.003$		&same	 &31.0 	&48.7&56.2\\
$0.01$		&diff.	 &31.0 	&\textbf{49.2}&56.3\\
$0.03$		&diff.	 &28.1\	&46.4&53.4\\
\hline
\end{tabular}
\end{center}
\caption{Influence of the balancing parameters.}
\label{table:balancing}
\end{table}

\begin{table}[t]
\begin{center}
\footnotesize
\begin{tabular}{c|ccc|c}
\hline
$r$&AP (\%)&PQ$^{\text{Th}}$ (\%)&PQ (\%)&mIoU (\%)\\
\hline
0&-&-&-&74.5\\
3&30.5&48.5&56.4&75.0\\
5&\textbf{31.3}&\textbf{49.0}&\textbf{56.5}&75.0\\
7&31.2&48.1&56.0&\textbf{75.1}\\
9&30.0&46.2&55.0&74.3\\
\hline
\end{tabular}
\end{center}
\caption{Influence of affinity window size $r$. mIoU for semantic segmentation evaluation is also provided. $r=0$ means to train semantic segmentation only.}
\label{table:windowsize}
\end{table}

\begin{table}[t]\addtolength{\tabcolsep}{-.5pt}
\begin{center}
\footnotesize
\begin{tabular}{cc|ccc|c}
\hline
Feature&JL&AP (\%)&PQ$^{\text{Th}}$ (\%)&PQ (\%)&mIoU (\%)\\
\hline
Single&&29.4&46.9&54.9&74.5\\
(w/o dilation)&$\surd$&30.2&47.6&55.0&74.2\\
\hline
Single&&30.6&48.2&55.5&74.5\\
(w/ dilation)&$\surd$&30.8&48.8&55.8&74.5\\
\hline
\multirow{2}*{Hierarchical}&&30.0&47.7&55.2&74.5\\
~&$\surd$&\textbf{31.3}&\textbf{49.0}&\textbf{56.5}&\textbf{75.0}\\
\hline
\end{tabular}
\end{center}
\caption{\textbf{JL}: joint learning. Comparing with learning all layers of the affinity pyramid from the single 1/4 resolution feature map, our hierarchical manner with joint learning performs better.}
\label{table:ablation_feature}
\end{table}

\noindent\textbf{Implementation Details }
Our model predicts semantic segmentation and pixel-pair affinity with a unified U-shape framework based on ResNet-50 \cite{he2016deep}. The training loss $L$ is defined as: 
\begin{equation}
\vspace{-.0em}
\label{eq:lossall}
L=\sum_i({L^i_s}+\alpha\lambda_i L^i_a),
\vspace{-.0em}
\end{equation}
where $L^i_s$ and $L^i_a$ are multi-class focal loss \cite{lin2017focal} and average L2 loss (see Eq.~\ref{eq:al2}) for semantic segmentation and affinity branches at the $i_{th}$ resolution in $[\frac{1}{4}, \frac{1}{8},..., \frac{1}{64}]$ resolutions respectively.
To combine losses from each scale, we firstly tune the balancing parameter $\lambda_i$ to make losses of each scale are in the same order, which are finally set to $[0.01, 0.03, 0.1, 0.3, 1]$ respectively. After that, $\alpha$ is set to 0.003 to balance the losses of affinity pyramid and semantic segmentation. The influence of $\alpha$ and $\lambda_i$ are shown in Table~\ref{table:balancing}.
We run all experiments using the MXNet framework \cite{chen2015mxnet}.
Our model is trained with Nadam \cite{dozat2016incorporating} for 70,000 iterations using synchronized batch normalization \cite{ioffe2015batch} over 8 TitanX 1080ti GPUs
and the batch size is set to 24. The learning rate is initialized to $10^{-4}$ and divided by 10 at the 30,000 and 50,000 iterations, respectively.

\noindent\textbf{Influence of Joint Learning }
Our separately trained semantic segmentation model achieves 74.5\% mIoU. This result is significantly improved after being jointly trained with the affinity pyramid, as shown in Table~\ref{table:windowsize}. However, the performance for both instance and semantic segmentation is affected by the affinity window size. 
Similar phenomenon is also observed by Ke \etal \cite{Ke_2018_ECCV} and they explain that small windows and large windows benefit small objects and large objects, respectively.
Due to the limitation of GPU memory, the window size is tested from 3 to 9. 
Among them, $5\times5$ affinity window balances the conflict and achieves the best performance, which is used in the other experiments.
Furthermore, in our proposed model, the semantic segmentation and affinity pyramid are jointly learned along the hierarchy of the U-shape network. We compare this approach with generating all layers of the affinity pyramid from the single $\frac{1}{4}$ resolution feature map with corresponding strides. The employing of dilated convolution ~\cite{chen2018deeplabv2:} is also tested.
Table~\ref{table:ablation_feature} shows our approach performs best,
where the mutually benefits of the two tasks are explored and finally improve the performance on instance segmentation.

\begin{figure}[t]
\begin{center}
\includegraphics[width=0.85\linewidth]{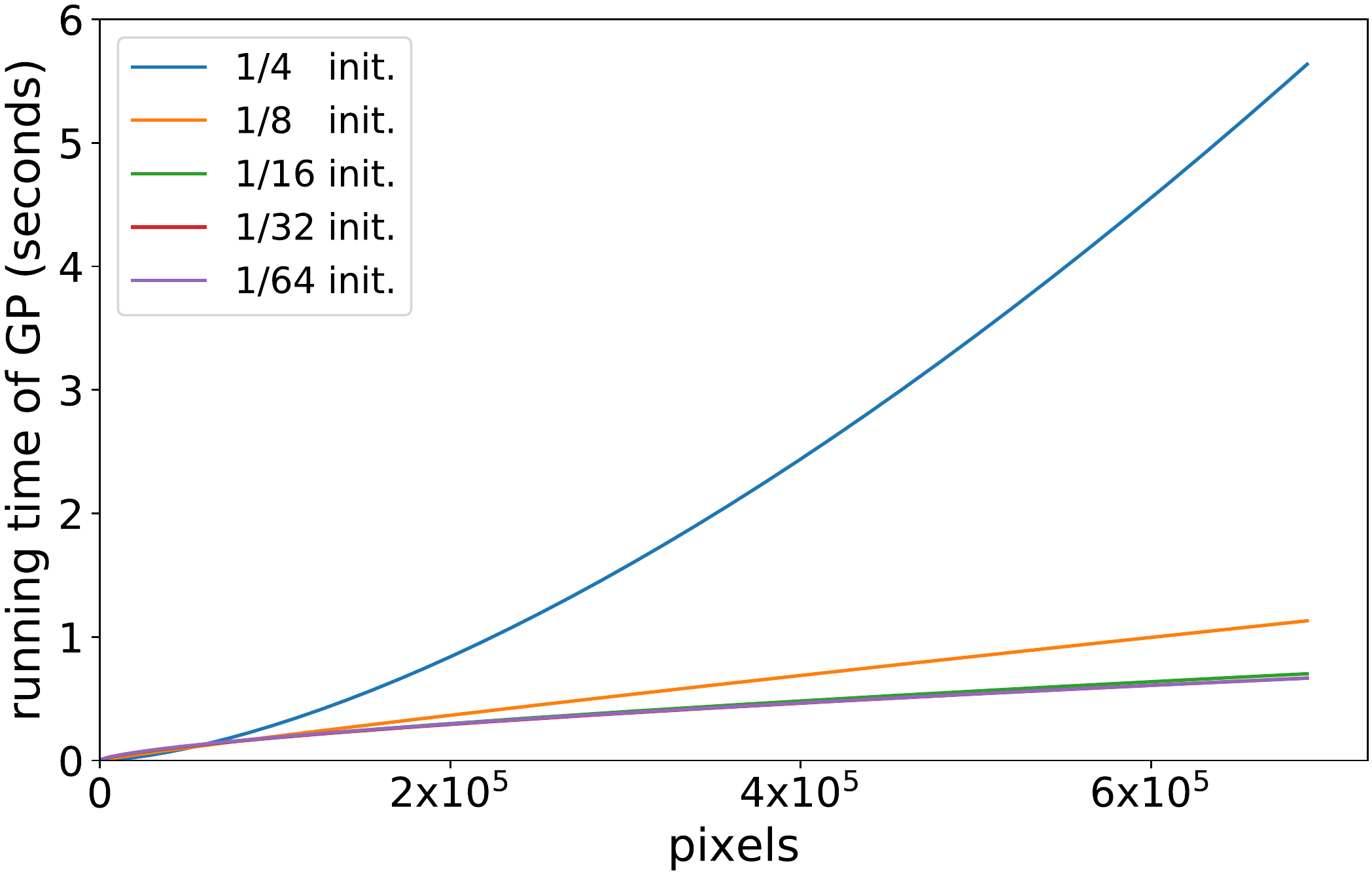}
\end{center}
\caption{Running time for the cascaded graph partition module under different object sizes. The cascade scheme significantly reduces the time for large objects. Best viewed in color.}
\label{fig:speed}
\end{figure}
\begin{table}[t]
\begin{center}
\footnotesize
\begin{tabular}{c|c|ccc}
\hline
Init. Res. & GP time (s)&AP (\%)&PQ$^{\text{Th}}$ (\%)&PQ (\%)\\
\hline
1/4  &1.26&28.9&45.1&54.9\\
1/8  &0.33&31.3&\textbf{49.2}&\textbf{56.6}\\
1/16 &0.26&\textbf{31.5}&\textbf{49.2}&\textbf{56.6}\\
1/32 &0.26&30.9&48.8&56.5\\
1/64 &0.26&30.9&48.7&56.5\\
\hline
\end{tabular}
\end{center}
\caption{Influence of the initial resolution for the cascaded graph partition. With the decreasing of initial resolution, the GP time (running time for cascaded graph partition per image) keeps decreasing. Comparing with $\frac{1}{4}$ resolution initialization, initializing cascaded graph partition from the $\frac{1}{16}$ resolution achieves $5\times$ speedup with 9\% AP improvement.}
\label{table:speedup}
\end{table}
\noindent\textbf{Influence of Cascaded Graph Partition }
At this part, the proposed cascaded graph partition module is analyzed by being initialized from each resolution.
As shown in Fig.~\ref{fig:speed}, the running time for graph partition increases rapidly \wrt the size of object regions when conducting the partition at the $\frac{1}{4}$ resolution directly, without the guidance of instance proposals.
However, the time significantly reduces when initializing the cascaded graph partition from lower resolutions, like the $\frac{1}{16}$ resolution, where the graph partition is constructed at the $[\frac{1}{16},\frac{1}{8},\frac{1}{4}]$ resolutions sequentially, and the latter two are guided by the proposals from the previous stage. 
The quantitative results are shown in Table~\ref{table:speedup}. 
Comparing with the $\frac{1}{4}$ resolution initialization (non-cascaded), the $\frac{1}{64}$ resolution initializing scheme achieves $5\times$ acceleration. 
Importantly, the cascaded approach achieves speeding up without scarifying precisions. As shown in Table~\ref{table:speedup}, initializing from the $\frac{1}{64}$ resolution has 2.0\% absolute improvement on AP, which is achieved due to that the proposals from lower resolutions can reduce the disturbing information for prediction. Meanwhile, the $\frac{1}{16}$ resolution initializing approach achieves better performance than the $\frac{1}{64}$ and $\frac{1}{32}$ manner, which indicates proposals from too low resolutions still bring errors for prediction.
In the other experiments, cascaded graph partitions are initialized from the $\frac{1}{16}$ resolution.

\begin{table}[t]
\begin{center}
\footnotesize
\begin{tabular}{c|ccc}
\hline
Affinities Used&AP (\%)&PQ$^{\text{Th}}$ (\%)&PQ (\%)\\
\hline
$A_1$ only  &25.7&41.2&53.2\\
$\quad\; + A_2$ &29.8&46.5&55.4\\
$\quad\; + A_3$ &30.8&48.6&56.3\\
$\quad\; + A_4$ &31.4&49.2&56.5\\
$\quad\; + A_5$ &31.5&49.2&56.6\\
\hline
\end{tabular}
\end{center}
\caption{Effectiveness of the long-range affinities. $[A_1,A_2,...,A_5]$ are affinities of the $[\frac{1}{4},\frac{1}{8},...,\frac{1}{64}]$ resolutions respectively. Affinities with longer-range are gradually added.}
\label{table:long-range}
\end{table}
\begin{table}[t]
\begin{center}
\footnotesize
\begin{tabular}{ccc|ccc}
\hline
BD&OL&Kernel&AP (\%)&PQ$^{\text{Th}}$ (\%)&PQ (\%)\\
\hline
&&3&29.1&46.4&55.8\\
$\surd$&&3&30.0&48.8&56.0\\
$\surd$&$\surd$&3&31.3&49.0&56.5\\
$\surd$&$\surd$&5&31.5&49.2&56.6\\
\hline
\end{tabular}
\end{center}
\caption{
\textbf{BD}: balance the training data by randomly dropping 80\% pixels with all 1 ground truth affinities.
\textbf{OL}: set 3 times affinity loss for pixels belonging to object instances.
\textbf{Kernel}: kernel size.}
\label{table:ablation_train}
\end{table}

\noindent\textbf{Quantitative Results } Firstly, to show the effectiveness of the long-range affinities, we start with just using the affinities from the 1/4 resolution, and gradually add longer-range affinities. Results are shown in Table~\ref{table:long-range}. Then, the influences of balancing training data, setting larger affinity loss and employing a large kernel are evaluated and shown in Table~\ref{table:ablation_train}. 
After that, as shown in Table~\ref{table:ablation_post}, the segmentation refinement improves the performance with 2.8\% AP.
With test tricks, our model achieves 34.4\% AP and 58.4\% PQ on the validation set.
\begin{table}[t]\addtolength{\tabcolsep}{-1.pt}
\begin{center}
\footnotesize
\begin{tabular}{cccc|ccc}
\hline
Backbone&SR&HF&MS&AP (\%)&PQ$^{\text{Th}}$ (\%)&PQ (\%)\\
\hline
ResNet-50&		 &		  &			&28.7&45.4&55.1\\
ResNet-50&$\surd$&		  &			&31.5&49.2&56.6\\
ResNet-50&$\surd$&$\surd$ &			&32.8&50.4&57.6\\
ResNet-50&$\surd$&$\surd$ &$\surd$	&34.4&50.6&58.4\\
ResNet-101&$\surd$&$\surd$&$\surd$	&37.3&55.0&61.1\\
\hline
\end{tabular}
\end{center}
\caption{
\textbf{SR}: segmentation refinement. 
\textbf{HF}: horizontal flipping test.
\textbf{MS}: multiscale test.}
\label{table:ablation_post}
\end{table}
\begin{table}[t]\addtolength{\tabcolsep}{-2pt}
\begin{center}
\footnotesize
\begin{tabular}{l|ccc|l}
\hline
Method&AP (\%)&PQ$^{\text{Th}}$ (\%)&PQ (\%)&Backbone\\
\hline
Li \etal \cite{Li_2018_ECCV} &28.6&42.5&53.8&ResNet-101\\
SGN \cite{Liu_2017_ICCV}         &29.2&-&-&-\\
Mask R-CNN \cite{He_2017_ICCV}   &31.5&49.6\footnote{}&-&ResNet-50\\
GMIS \cite{Liu_2018_ECCV}  &34.1&-&-&ResNet-101\\
Deeperlab \cite{arxiv_2019_yang_deeperlab}&-&-&56.5&Xception-71 \cite{chollet2017xception:}\\
PANet \cite{Liu_2018_CVPR}       &36.5&-&-&ResNet-50\\
\hline
SSAP (ours)                        &34.4&50.6&58.4&ResNet-50\\
SSAP (ours)                        &\textbf{37.3}&\textbf{55.0}&\textbf{61.1}&ResNet-101\\
\hline
\end{tabular}
\end{center}
\caption{Results on Cityscapes \textbf{val} set. All results are trained with Cityscapes data only.}
\label{table:cityvalresult}
\end{table}
\begin{table}[t]
\addtolength{\tabcolsep}{-4.2pt}
\begin{center}
\scriptsize
\begin{tabular}{l|c|ccc|ccc|ccc}
\hline
Method&PQ [val]&PQ&SQ&RQ&PQ$^{\text{Th}}$ &SQ$^{\text{Th}}$&RQ$^{\text{Th}}$&PQ$^{\text{St}}$&SQ$^{\text{St}}$&RQ$^{\text{St}}$\\
\hline
DeeperLab \cite{arxiv_2019_yang_deeperlab} &33.8&34.3&77.1&43.1&37.5&77.5&46.8&29.6&76.4&37.4\\
\hline
SSAP (ours) &36.5&36.9&80.7&44.8&40.1&81.6&48.5&32.0&79.4&39.3\\
\hline
\end{tabular}
\end{center}
\vspace{-.0em}
\caption{{Results on COCO \textbf{val} (`PQ [val]' column ) and  \textbf{test-dev} (remaining columns) sets. Results are reported as percentages.}}
\vspace{-.0em}
\label{table:cocoresult}
\end{table}
\footnotetext[1]{This result is reported by Kirillov \etal \cite{panoptic}.}
Our model is also trained with ResNet-101, which achieves 37.3\% AP and 61.1\% PQ, as shown in Table~\ref{table:cityvalresult}.
For the test set, our model attains a performance of 32.7\% AP,
which exceeds all previous methods. Details are in Table~\ref{table:citytestresult}.

\begin{figure*}
\begin{center}
  \begin{tabular}{ccccc}
    \includegraphics[width=0.165\linewidth]{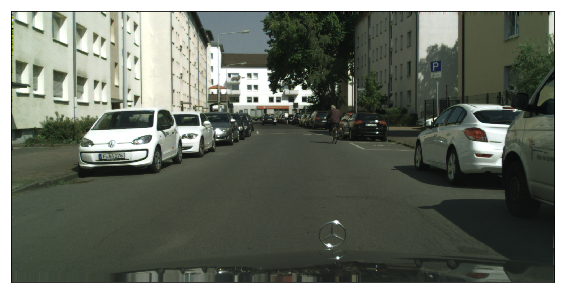}&
    \includegraphics[width=0.165\linewidth]{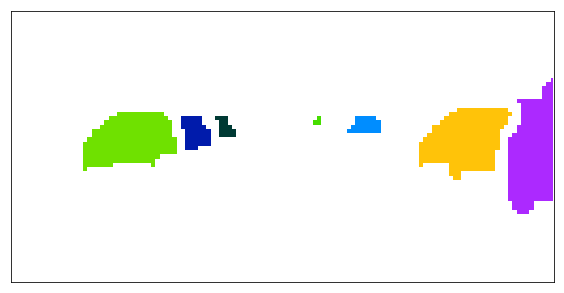}&
    \includegraphics[width=0.165\linewidth]{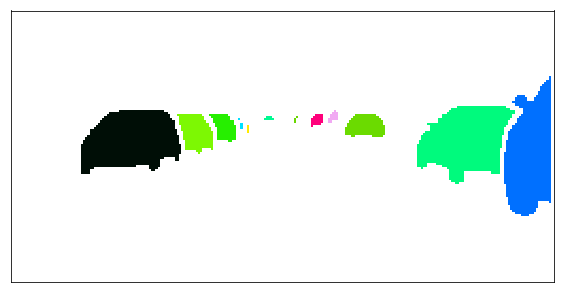}&
    \includegraphics[width=0.165\linewidth]{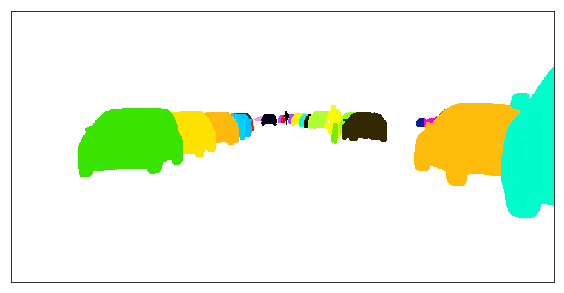}&
    \includegraphics[width=0.165\linewidth]{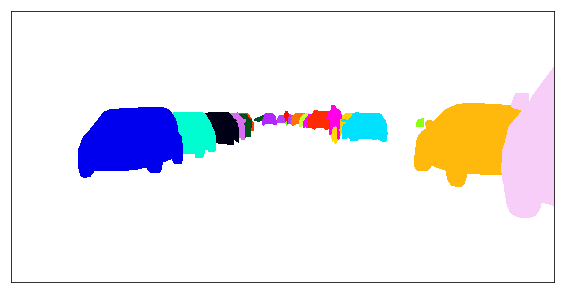}\\
    \includegraphics[width=0.165\linewidth]{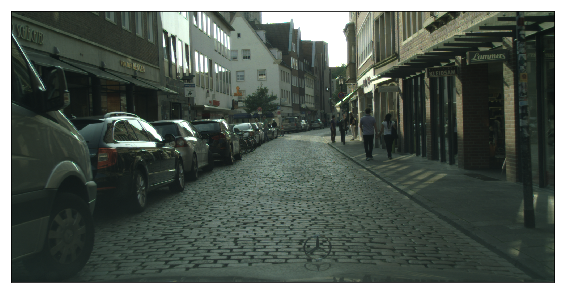}&
    \includegraphics[width=0.165\linewidth]{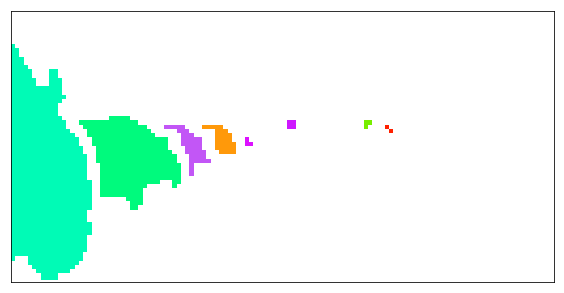}&
    \includegraphics[width=0.165\linewidth]{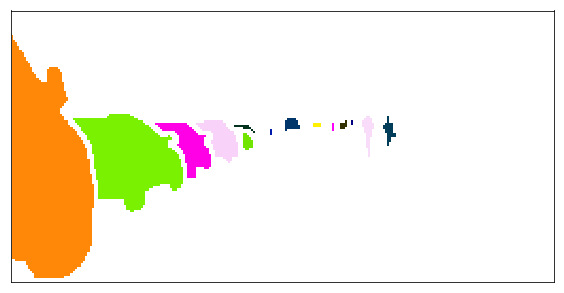}&
    \includegraphics[width=0.165\linewidth]{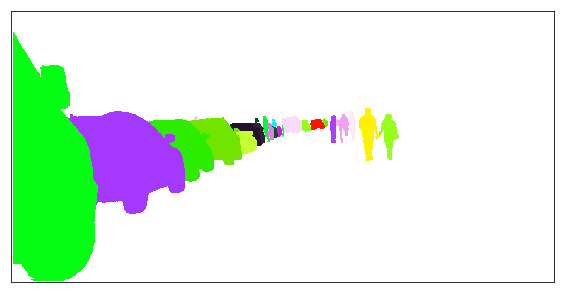}&
    \includegraphics[width=0.165\linewidth]{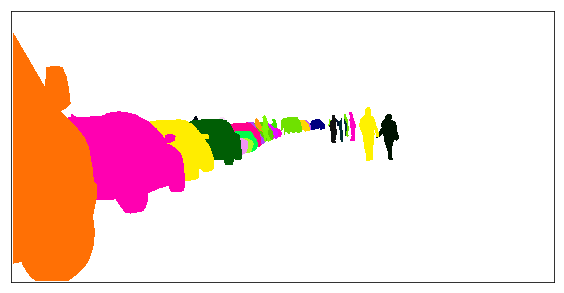}\\
    Image&Proposals from $\frac{1}{16}$ Res.&Proposals from $\frac{1}{8}$ Res.&Instance Seg.&Ground Truth\\
  \end{tabular}
\end{center}
  \caption{Visualizations of  proposals generated from lower resolutions within the cascaded graph partition module and the final instance segmentation results. Best viewed in color and zoom.}
  \vspace{-0.em}
\label{fig:proposals}
\begin{center}
  \begin{tabular}{cccc}
    \includegraphics[width=0.22\linewidth]{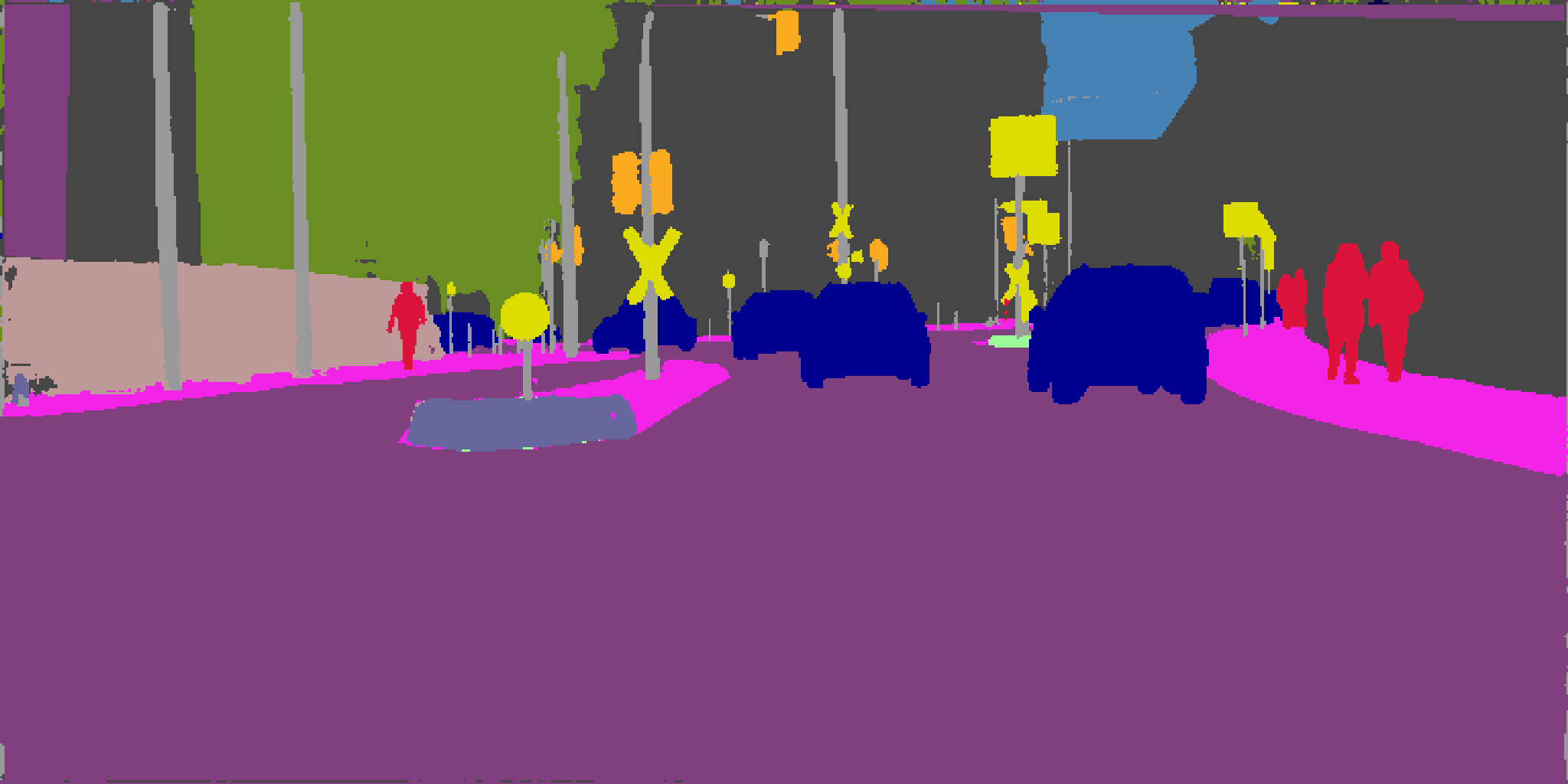}&
    \includegraphics[width=0.22\linewidth]{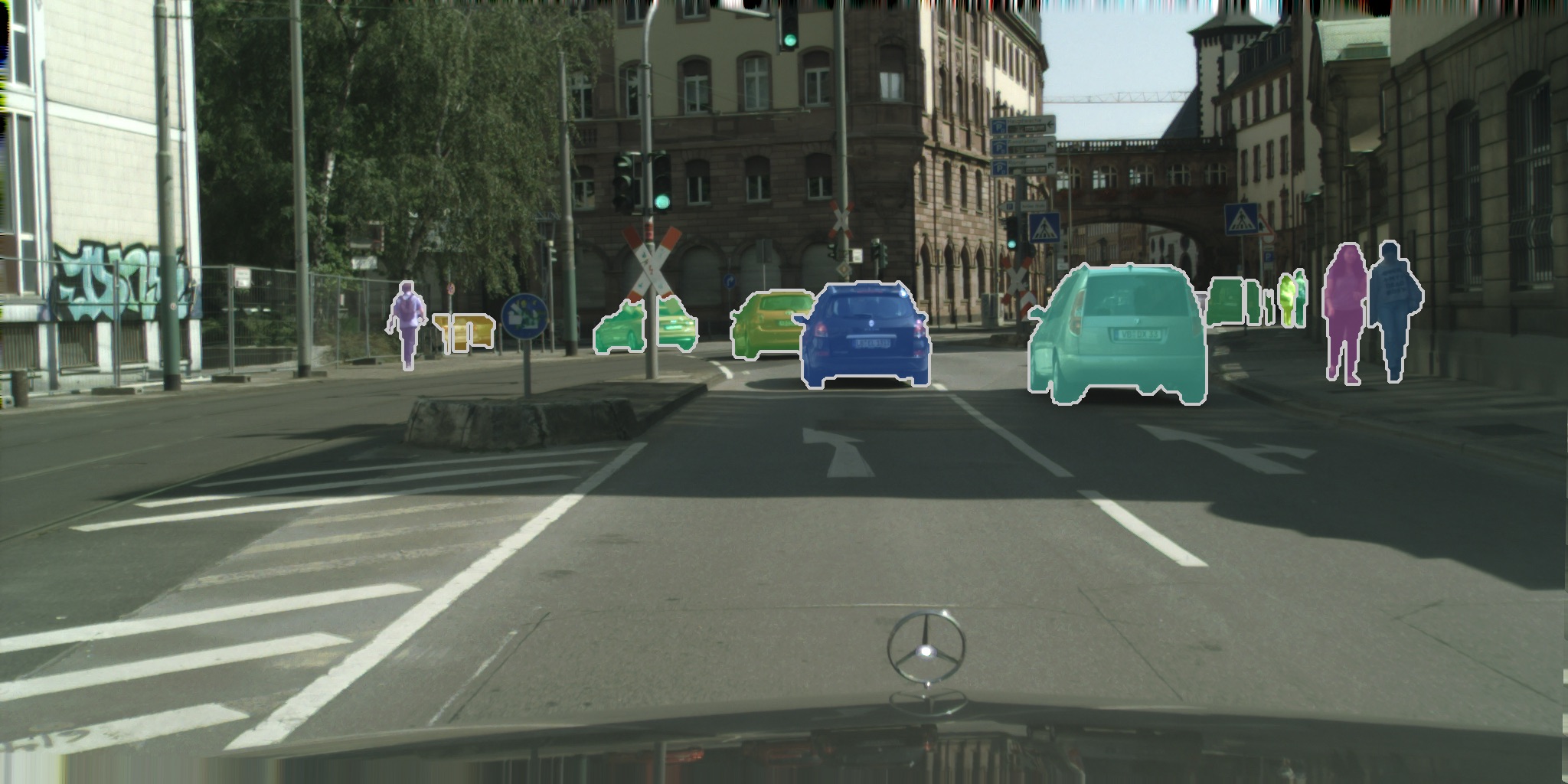}&
    \includegraphics[width=0.22\linewidth]{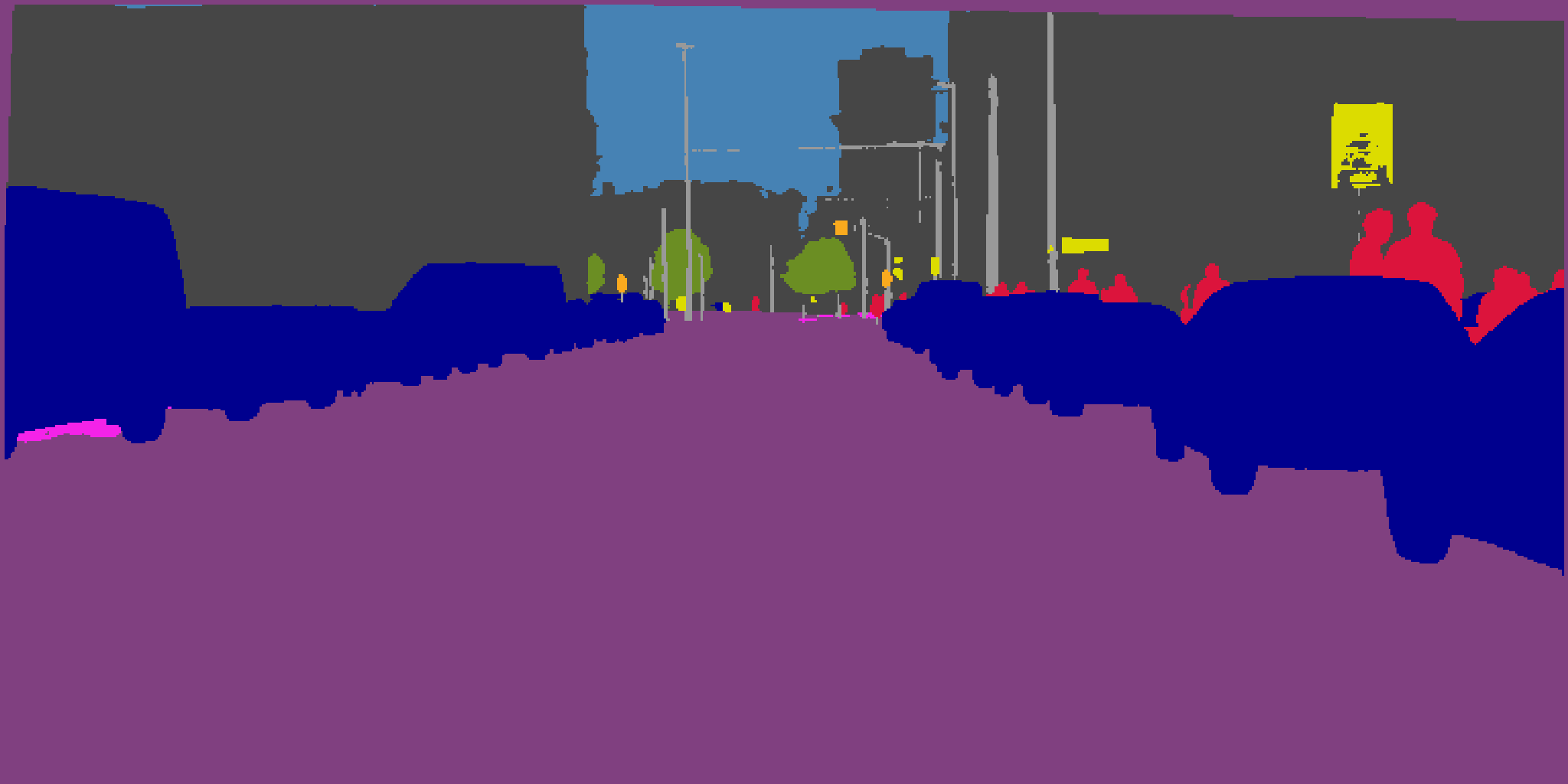}&
    \includegraphics[width=0.22\linewidth]{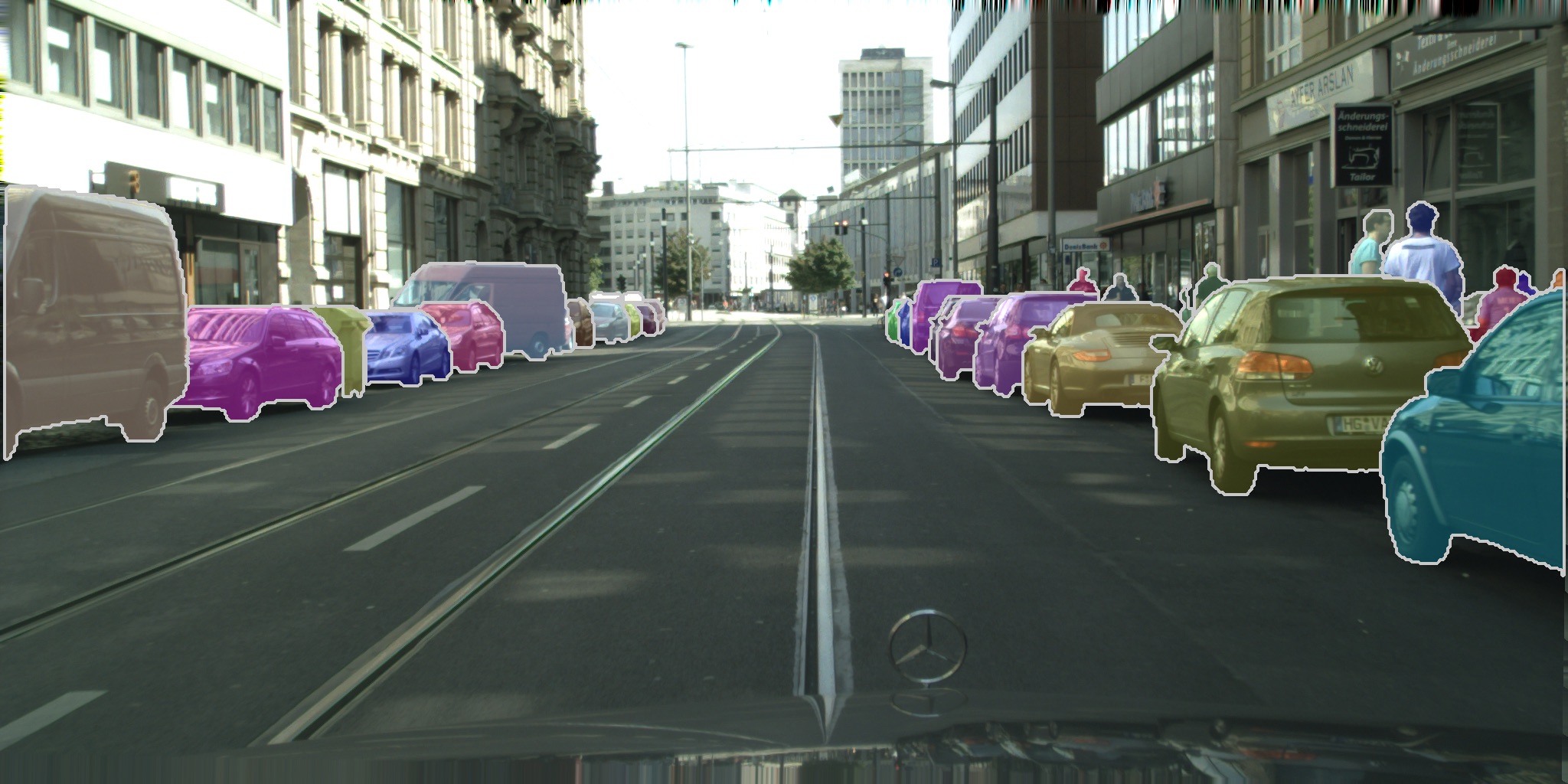}\\
    \includegraphics[width=0.22\linewidth]{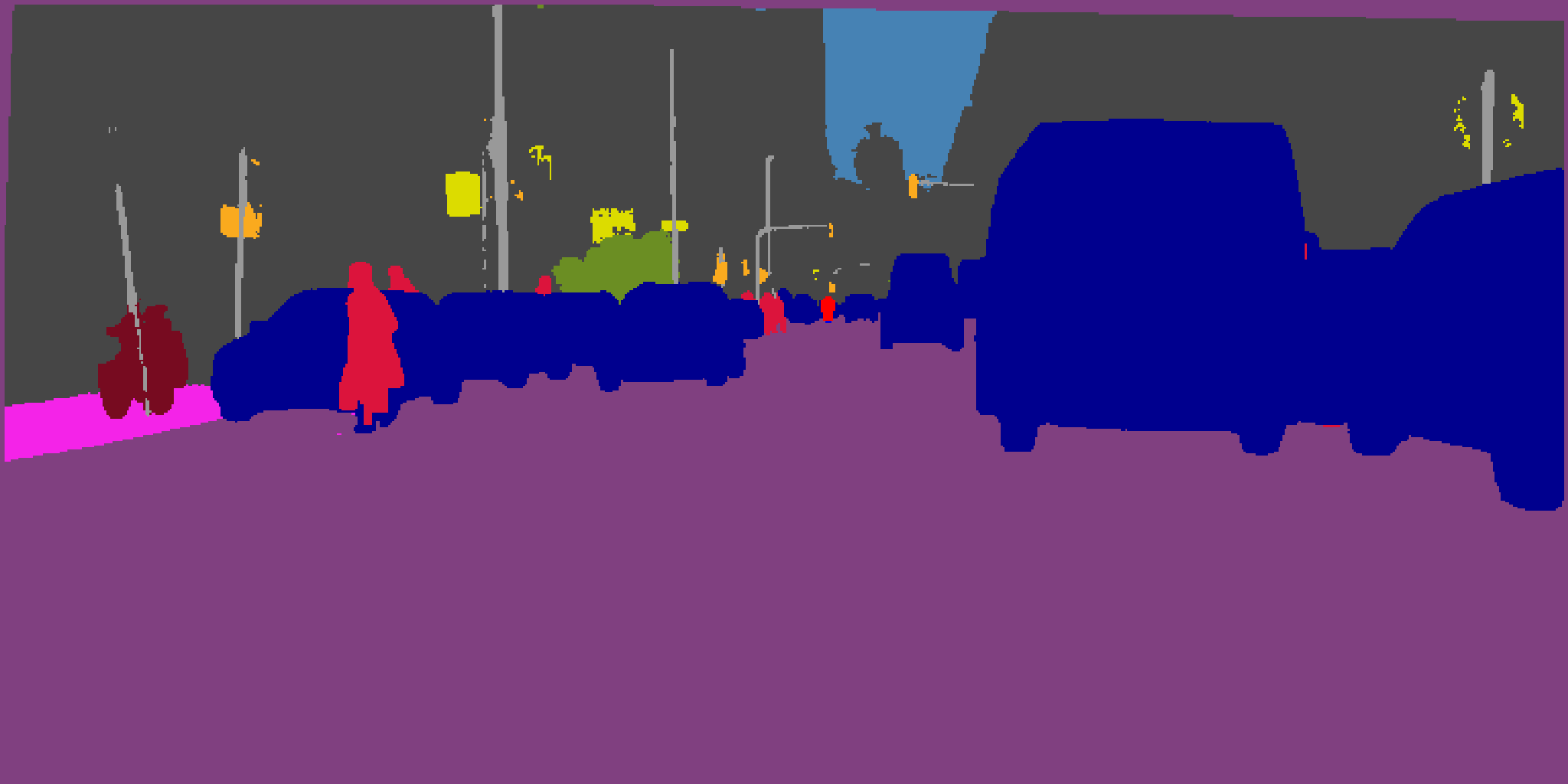}&
    \includegraphics[width=0.22\linewidth]{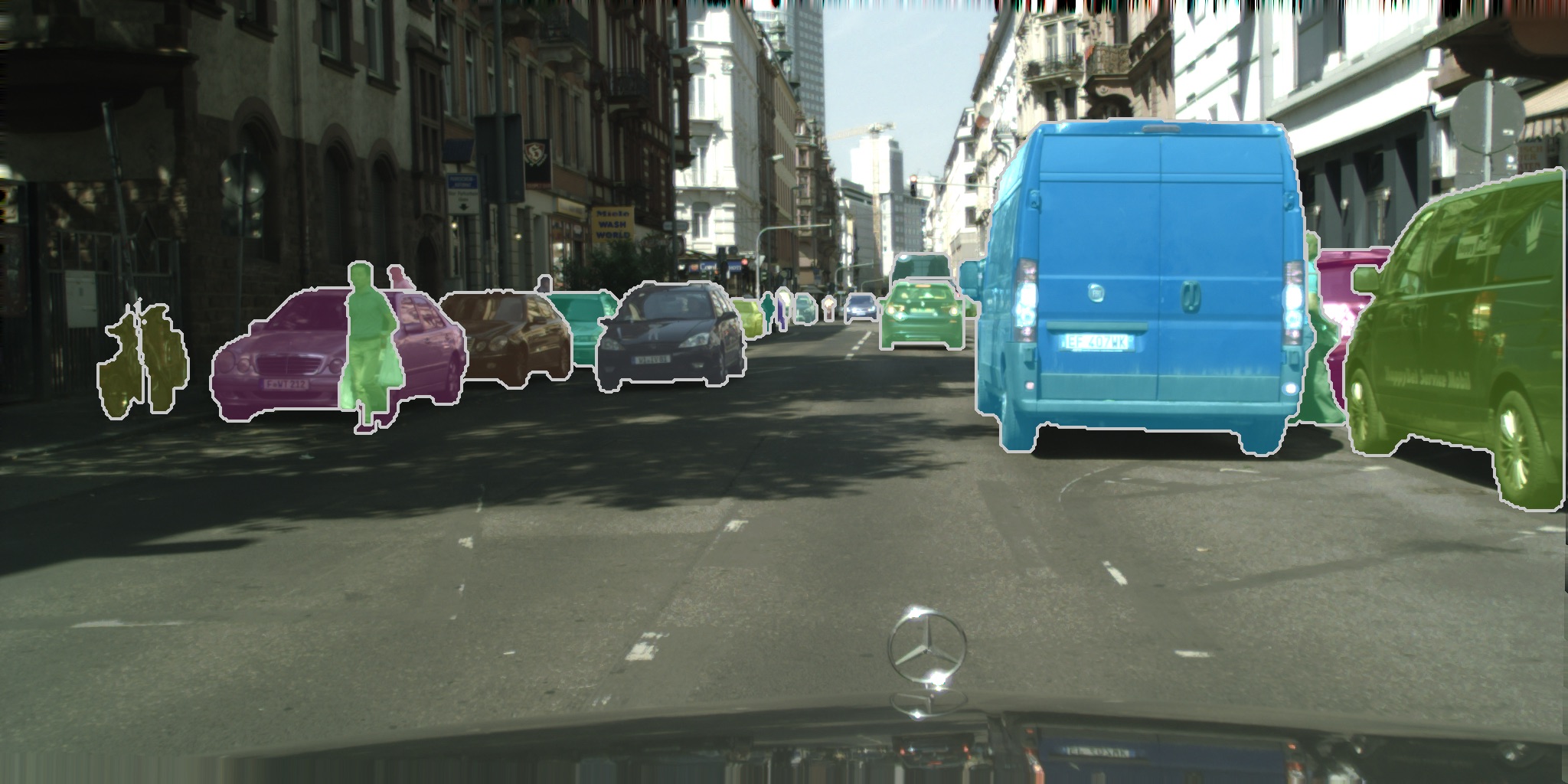}&
    \includegraphics[width=0.22\linewidth]{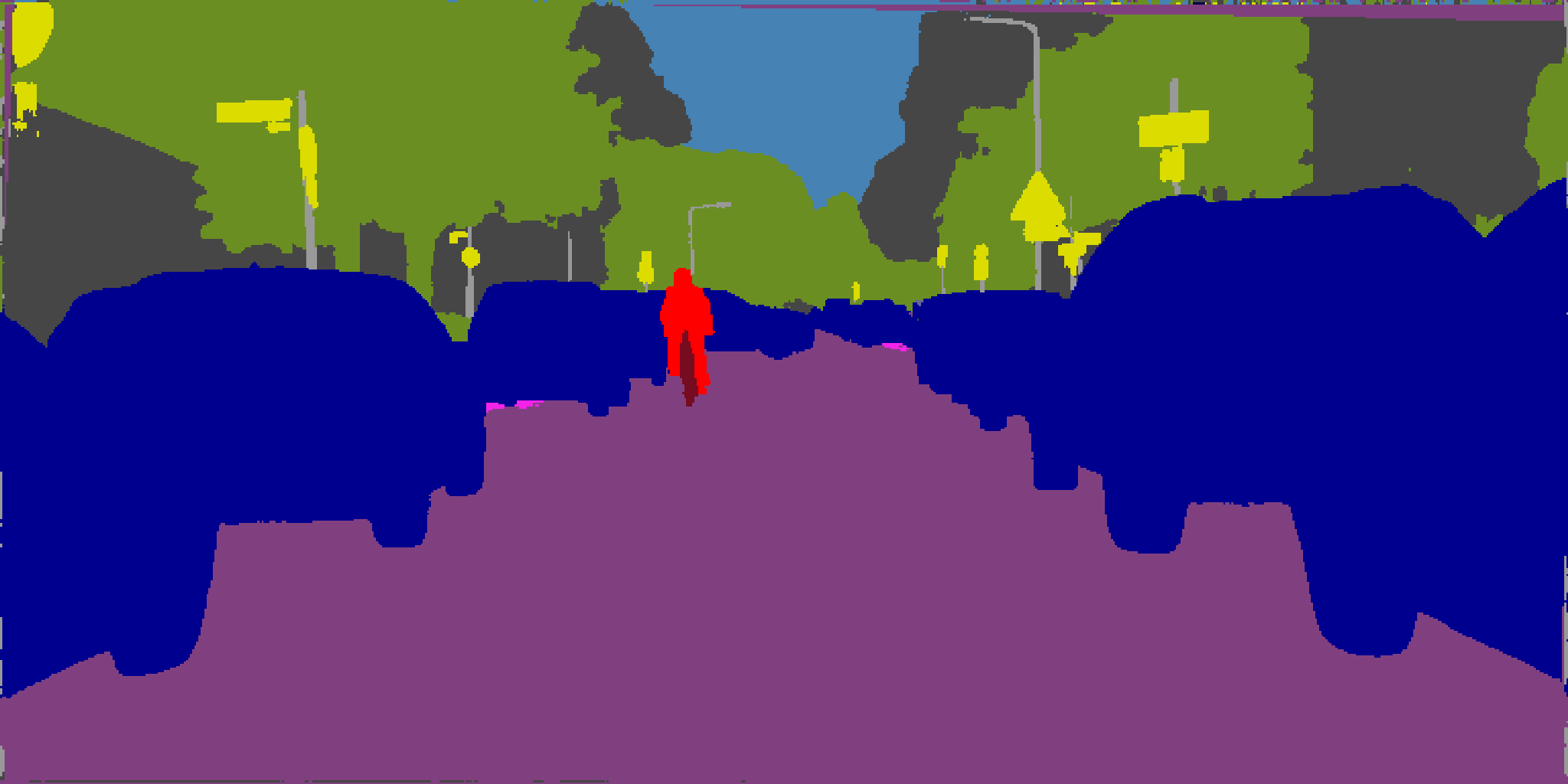}&
    \includegraphics[width=0.22\linewidth]{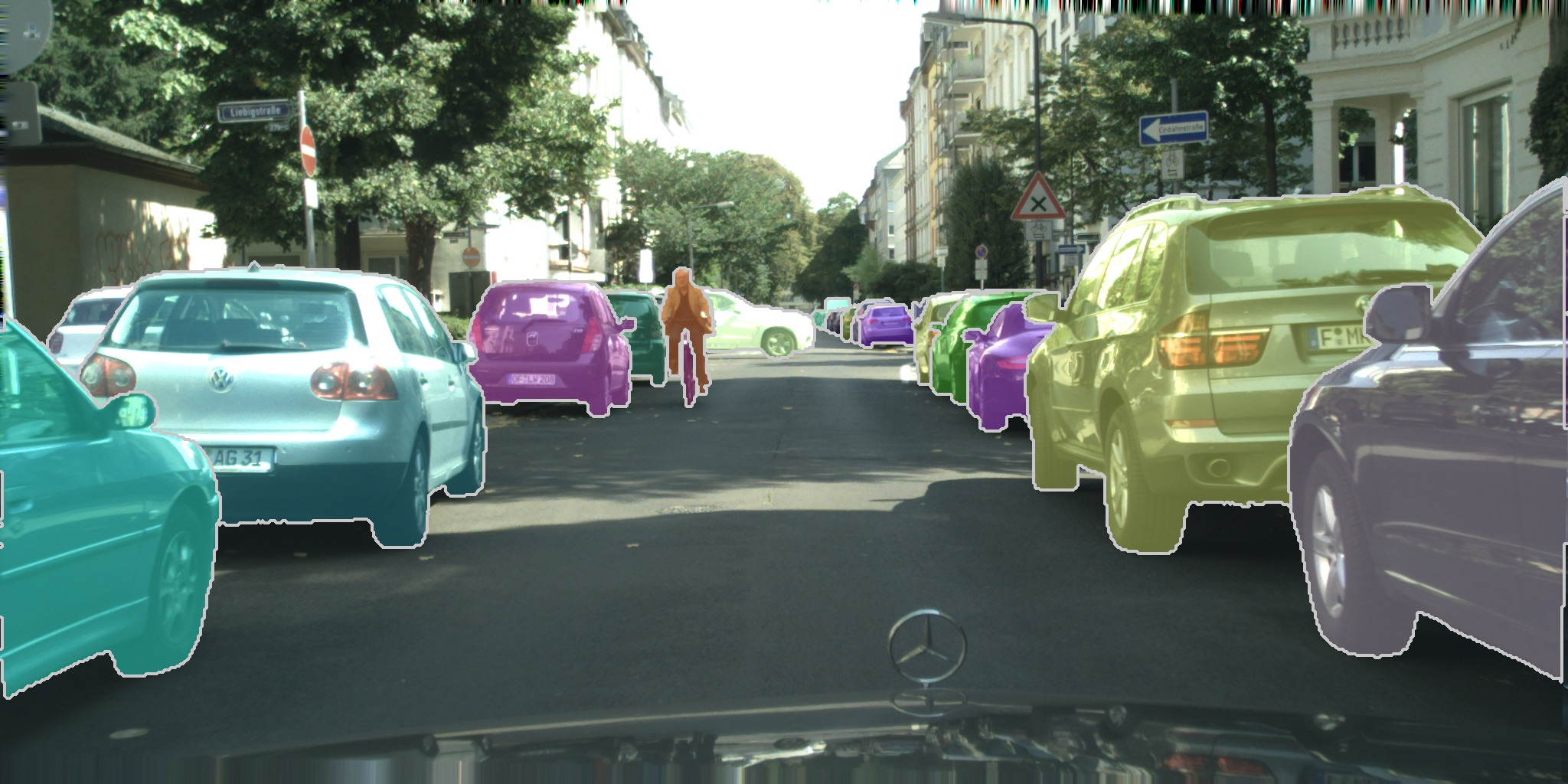}\\
    \includegraphics[width=0.22\linewidth]{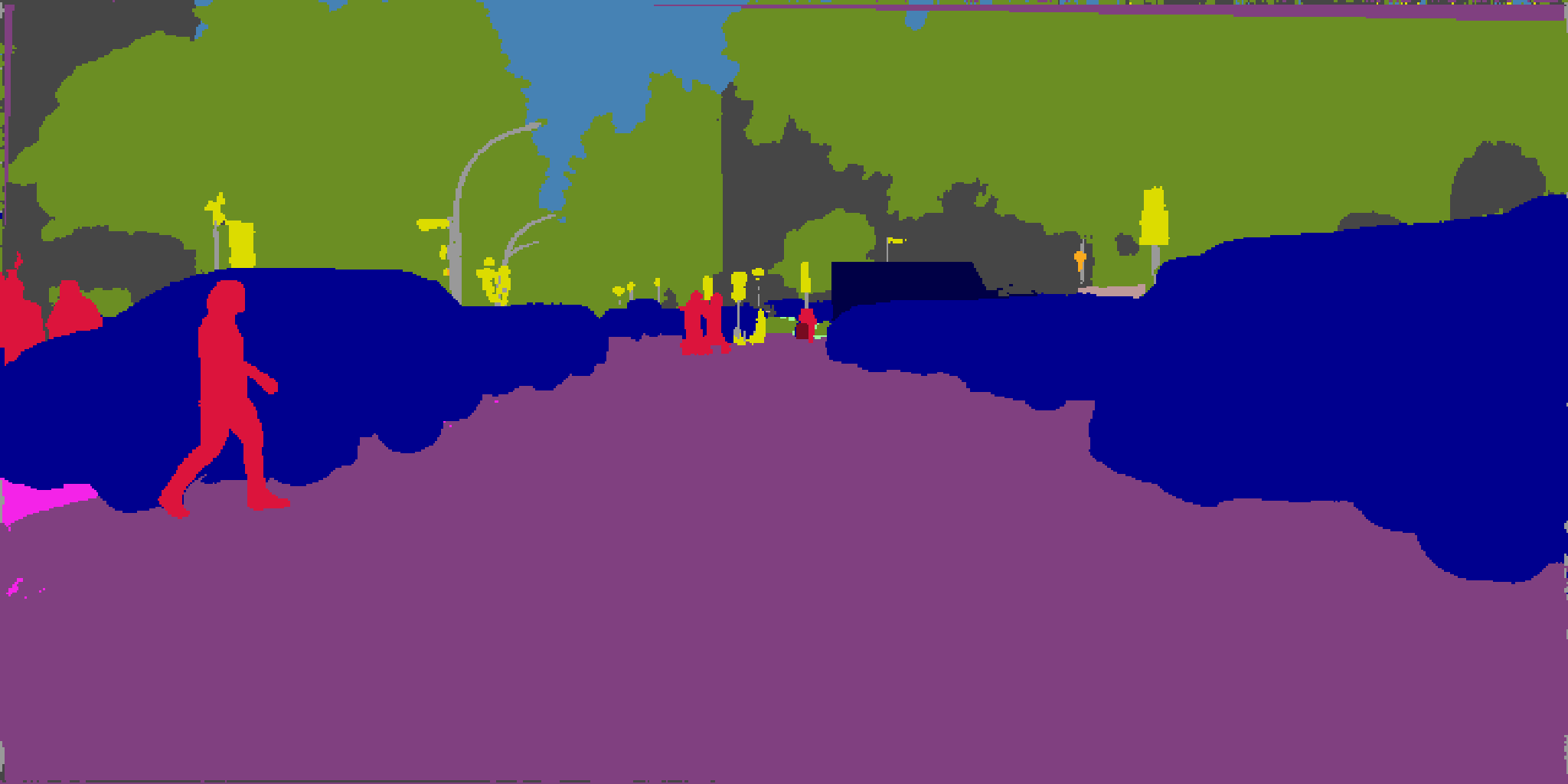}&
    \includegraphics[width=0.22\linewidth]{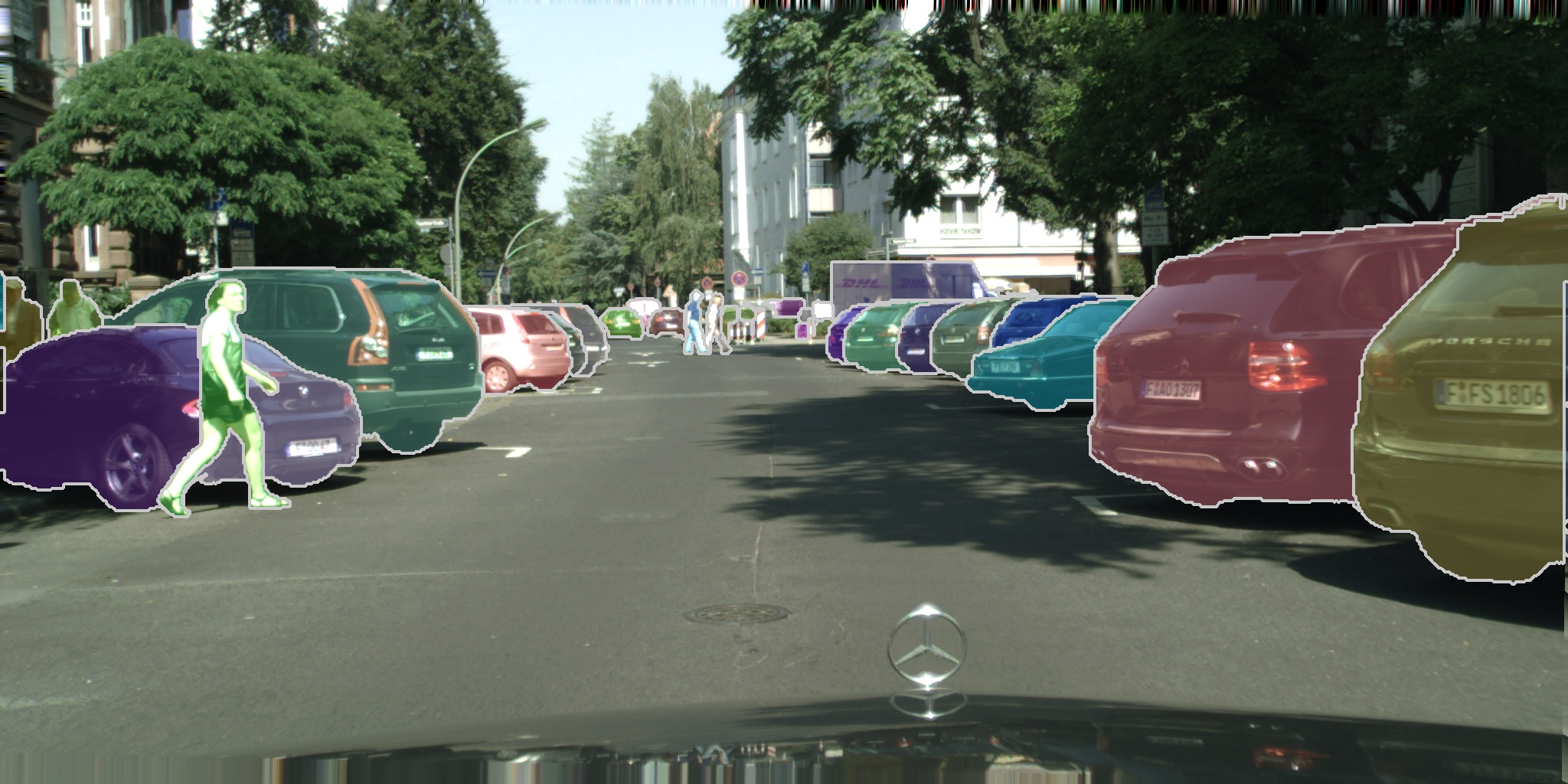}&
    \includegraphics[width=0.22\linewidth]{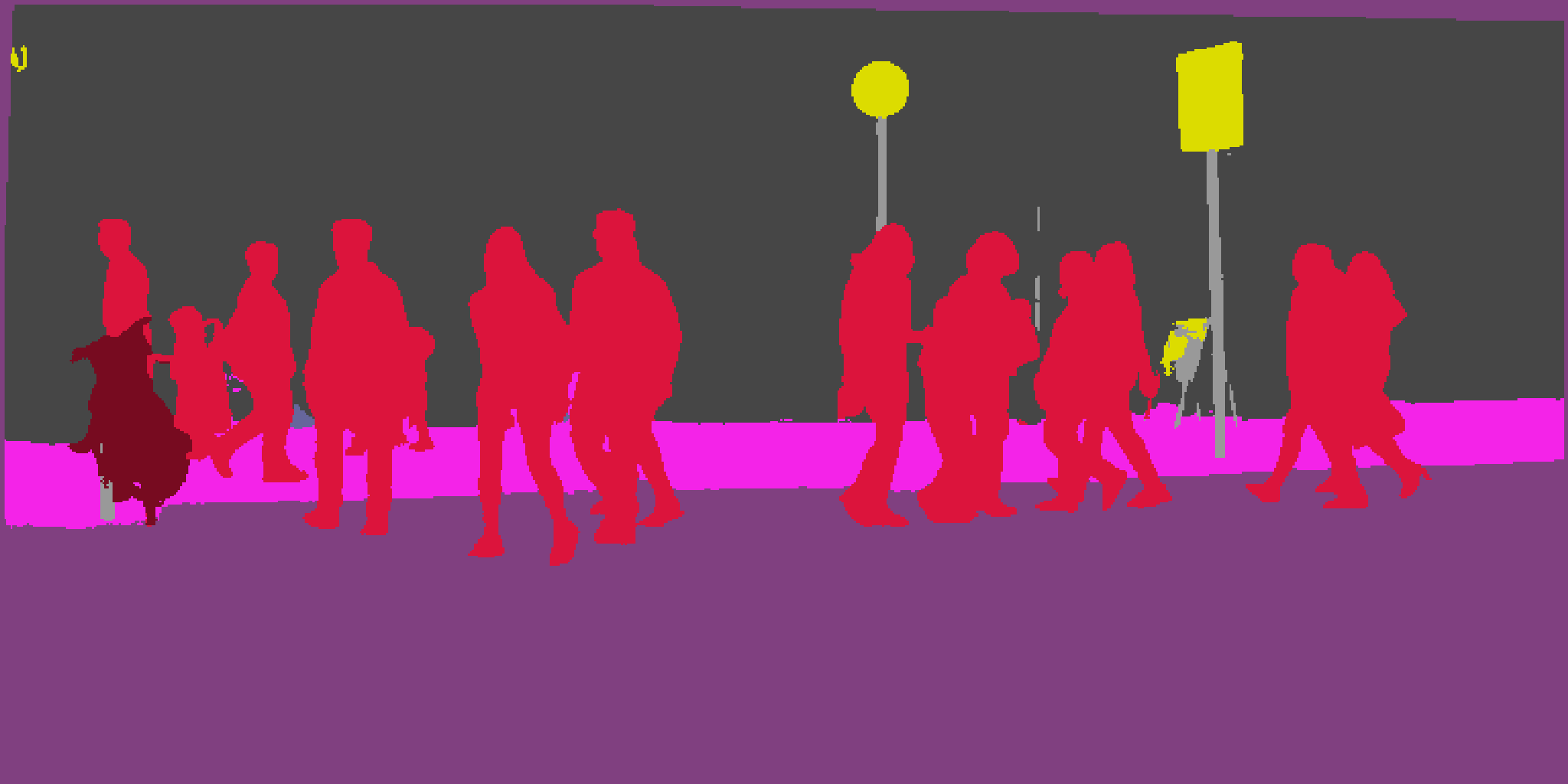}&
    \includegraphics[width=0.22\linewidth]{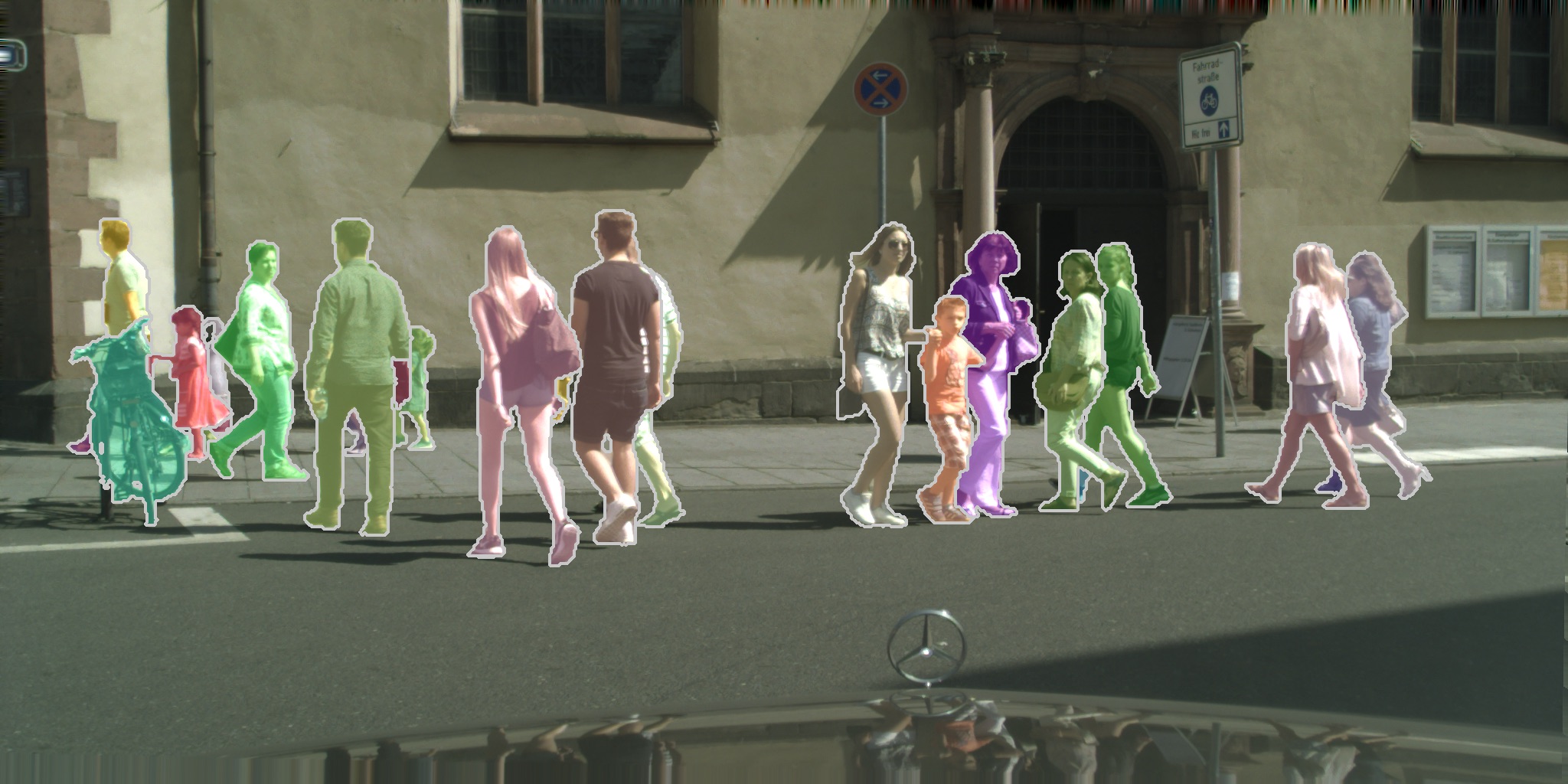}\\
    Semantic Seg.&Instance Seg.&Semantic Seg.&Instance Seg.\\
  \end{tabular}
\end{center}
  \caption{Visualizations of sampled results on the validation set. Best viewed in color and zoom.}
\label{fig:visualization}
\end{figure*}

\noindent\textbf{Visual Results }
The proposals generated from the $\frac{1}{16}$ and $\frac{1}{8}$ resolutions are visualized in Fig~\ref{fig:proposals}. A few sample results on the validation set are visualized in Fig~\ref{fig:visualization}, where fine details are precisely captured.
As shown in the second column, the cars occluded by persons or poles and separated into parts are successfully grouped. 

\noindent\textbf{Results on COCO }
To show the effectiveness of our method in scenarios other than streets, we evaluate it on the COCO dataset. The annotations for COCO instance segmentation are with overlaps, making it unsuitable to train and test a proposal-free method like ours. So our method is evaluated in the panoptic segmentation task. 
To train on COCO, we resize the longer edge to 640 and train the model with $512\times512$ crops. The number of iterations is 80,000 and the learning rate is divided by 10 in 60,000 and 70,000 iterations. Other experimental settings are remained the same.
The performance of our model (ResNet-101 based) is summarized in Table ~\ref{table:cocoresult}.
To the best of our knowledge, DeeperLab \cite{arxiv_2019_yang_deeperlab} is currently the only proposal-free method to report COCO result.
Our method outperformes DeeperLab (Xception-71 based) in all sub metrics.

\begin{table*}[htbp]
\addtolength{\tabcolsep}{-1.pt}
\begin{center}
\footnotesize
\begin{tabular}{l|l|cccc|cccccccc}
\hline
Method&{Training data}&AP&AP50\%&AP50m&AP100m&{person}&{rider}&car&{trunk}&{bus}&{train}&{motor}&{bicycle}  \\
\hline
InstanceCut \cite{Kirillov_2017_CVPR} &\tt{fine$\,$+$\,$coarse}&13.0&27.9&26.1&22.1&10.0&8.0&23.7&14.0&19.5&15.2&9.3&4.7\\
Multi-task \cite{kendall2018multi-task}&\tt{fine}&21.6&39.0&37.0&35.0&19.2&21.4&36.6&18.8&26.8&15.9&19.4&14.5\\
SGN \cite{Liu_2017_ICCV}&\tt{fine$\,$+$\,$coarse}&25.0&44.9&44.5&38.9&21.8&20.1&39.4&24.8&33.2&30.8&17.7&12.4\\
Mask RCNN  \cite{He_2017_ICCV}&\tt{fine}&26.2&49.9&40.1&37.6&30.5&23.7&46.9&22.8&32.2&18.6&19.1&16.0\\
GMIS \cite{Liu_2018_ECCV}&\tt{fine$\,$+$\,$coarse}&27.3&45.6&-&-&31.5&25.2&42.3&21.8&37.2&28.9&18.8&12.8\\
Neven \etal \cite{Neven_2019_CVPR} &\tt{fine}&27.6&50.9&-&-&34.5&26.1&52.4&21.7&31.2&16.4&20.1&18.9\\
PANet  \cite{Liu_2018_CVPR} &\tt{fine}&31.8&\textbf{57.1}&46.0&44.2&\textbf{36.8} &\textbf{30.4} &54.8 &27.0 &36.3 &25.5 &\textbf{22.6} &\textbf{20.8}\\
\hline
SSAP (ours) &\tt{fine}&\textbf{32.7}&51.8&\textbf{51.4}&\textbf{47.3}&35.4&25.5&\textbf{55.9}&\textbf{33.2}&\textbf{43.9}&\textbf{31.9}&19.5&16.2\\
\hline
\end{tabular}
\end{center}
\caption{Results on Cityscapes \textbf{test} set. All results are trained with Cityscapes data only. Results are reported as percentages.}
\label{table:citytestresult}
\end{table*}

\section{Conclusion}
This work has proposed a single-shot proposal-free instance segmentation method, which requires only one single pass to generate instances. Our method is based on a novel affinity pyramid to distinguish instances, which can be jointly learned with the pixel-level semantic class labels using a single backbone network. 
Experiment results have shown the two sub-tasks are mutually benefited from our joint learning scheme, which further boosts instance segmentation.
Moreover, a cascaded graph partition module has been developed to segment instances with the affinity pyramid and semantic segmentation results.
Comparing with the non-cascaded way, this module has achieved $5\times$ speedup and 9\% relative improvement on AP. Our approach has achieved a new state of the art on the challenging Cityscapes dataset.

\section*{Acknowledgment}
This work is supported in part by the National Key Research and Development Program of China (Grant No.2016YFB1001005), the National Natural Science Foundation of China (Grant No. 61673375 and No.61602485), and the Projects of Chinese Academy of Science (Grant No. QYZDB-SSW-JSC006).

{\small
\bibliographystyle{ieee_fullname}
\bibliography{SSAP}
}

\end{document}